%% file: main.tex
\documentclass{article} 
\usepackage{iclr2027_conference,times}

\input{math_commands.tex}

\usepackage{hyperref}
\usepackage{cleveref}
\usepackage{float}
\usepackage{algorithm}
\usepackage{algorithmic}
\usepackage{amssymb}
\usepackage{pifont}
\usepackage{enumitem}
\usepackage{mathtools}
\usepackage{url}
\usepackage{minitoc} 
\usepackage{graphicx} 
\usepackage{caption} 
\usepackage{subcaption}

\crefname{subfigure}{figure}{figures}
\Crefname{subfigure}{Figure}{Figures}
\usepackage{wrapfig} 
\usepackage{multirow}  
\usepackage{array}     
\usepackage{booktabs}  
\usepackage{tikz}
\usetikzlibrary{arrows.meta,positioning,calc,patterns,fit}
\usepackage{pgfplots}
\usepgfplotslibrary{groupplots}
\usepackage{xcolor}
\pgfplotsset{compat=1.16}
\input{figures/asap_style.tex}
\DeclareCaptionFont{asaptables}{\fontsize{9}{10.5}\selectfont}
\DeclareCaptionFont{asappanel}{\figfont{8.5}}

\title{ASAP: Arbitrary-Step Safe Planning with MeanFlow and Control Barrier Functions}

\author{\fontsize{9pt}{10.8pt}\selectfont 
Tzu-Yuan Huang$^{1,\dagger}$, 
Yi Yue$^{1}$, 
Dai-Jie Wu$^{2}$,
Stefan Sosnowski$^{1}$,
\\
\fontsize{9pt}{10.8pt}\selectfont \bf  
Shih-Hsin Wang$^{3}$,
Kenneth Marino$^{2}$, 
Armin Lederer$^{4}$,
Sandra Hirche$^{1,5,6}$
\\[0.8em]
\fontsize{9pt}{10.8pt}\selectfont
$^1$Technical University of Munich \\
$^2$University of Utah\\
$^3$National Taiwan University \\
$^4$National University of Singapore\\
$^5$Munich Institute of Robotics and Machine Intelligence (MIRMI) \\
$^6$Munich Data Science Institute (MDSI) \\
Correspondence to: \texttt{tzu-yuan.huang@tum.de}$^{\dagger}$.
}

\definecolor{ForestGreen}{RGB}{34,139,34}
\newcommand{\cmark}{\textcolor{ForestGreen}{\ding{51}}} 
\newcommand{\xmark}{\textcolor{red}{\ding{55}}}       

\iclrfinalcopy 
\begin{document}
\doparttoc 
\faketableofcontents 
\maketitle

\begin{abstract}
Generative models have emerged as a powerful paradigm for data-driven robotic planning. Existing flow-matching planners typically rely on multi-step sampling and lack principled mechanisms for enforcing safety constraints at test time. Bridging these limitations is essential for real-world deployment, where planners must support both fast generation and formal safety guarantees. We propose \textbf{A}rbitrary-Step \textbf{SA}fe \textbf{P}lanning (ASAP), a control-augmented MeanFlow framework for safe generation with any number of sampling steps, including one step. ASAP introduces a virtual control input governed by an affine barrier constraint that accounts for the sampling step size, yielding formal safety guarantees of generation for any number of steps. Without retraining, ASAP enforces unseen test-time constraints across navigation, locomotion, manipulation, and dexterous manipulation benchmarks, while planning at least $20\times$ faster than safety-guaranteed baselines on navigation, locomotion, and manipulation, and $2\times$ faster on dexterous manipulation. We further validate ASAP on a real-world robot handover task, where it maintains safety and high task performance under an obstacle, requiring only $0.04$\,s per plan and achieving over a $200\times$ speedup over the safe baseline.

\end{abstract}
\input{sections/introduction.tex}

\input{sections/related_work.tex}
\input{sections/problem_setting.tex}

\input{sections/method.tex}

\input{sections/experiments.tex}

\input{sections/conclusion.tex}


\subsection*{AI Use Statement}

Generative AI tools, including large language models such as ChatGPT, were used to assist with language editing, rephrasing, LATEX formatting, and code implementation during manuscript preparation. They were not used to develop the research ideas, theoretical results, or core experimental methodology. All AI-assisted work was reviewed and revised by the authors, who take full responsibility for the final content of the paper.

\subsection*{Reproducibility Statement}

\textbf{Code.}
Our implementation builds on the publicly available codebases of Flow Guidance~\citep{feng2025on} (\url{https://github.com/AI4Science-WestlakeU/flow_guidance}), SAD-Flower~\citep{huang2026sad} (\url{https://github.com/TzuYuan107/sad-flower}), MeanFlow~\citep{geng2025mean} (\url{https://github.com/Gsunshine/meanflow}), and MP1~\citep{sheng2026mp1} (\url{https://github.com/LogSSim/MP1}). The code for ASAP will be released upon acceptance. Implementation and experimental details are provided in \Cref{sec:exp}, with additional specifications in the Appendix.

\textbf{Theory.}
All theoretical results presented in \Cref{sec:method} are accompanied by the proof in Appendix~\ref{append:proof}.

\textbf{Datasets.}
We evaluate ASAP on publicly available datasets spanning navigation, locomotion, robotic manipulation, and dexterous manipulation. Dataset sources and preprocessing details are described in \Cref{sec:exp} and Appendix~\ref{append:env}.

\textbf{Compute.}
All the training of each experiment are conducted using NVIDIA RTX 4090 GPUs. Detailed hardware configurations and computational resources are reported in Appendix~\ref{append:compute}.


\bibliography{iclr2027_conference}
\bibliographystyle{iclr2027_conference}

\input{sections/appendix}

\end{document}

%% file: math_commands.tex
\usepackage{amsmath,amsfonts,bm,amsthm}

\def\eqref#1{equation~\ref{#1}}

\def\1{\bm{1}}

\DeclareMathAlphabet{\mathsfit}{\encodingdefault}{\sfdefault}{m}{sl}
\SetMathAlphabet{\mathsfit}{bold}{\encodingdefault}{\sfdefault}{bx}{n}

\DeclareMathOperator*{\argmin}{arg\,min}

\newtheorem{theorem}{Theorem}[section]
\newtheorem{lemma}{Lemma}[section]
\newtheorem{corollary}{Corollary}[section]
\newtheorem{proposition}{Proposition}[section]
\newtheorem{remark}{Remark}[section]


%% file: figures/asap_style.tex
\definecolor{fig_ink}{HTML}{1A1A1A}    
\definecolor{fig_muted}{HTML}{6E6E6E}  
\definecolor{fig_grid}{HTML}{E3E6EA}   
\definecolor{fig_edge}{HTML}{2B2B2B}   
\definecolor{fig_rule}{HTML}{9AA0A6}   

\definecolor{propose_color}{HTML}{EC4617} 
\definecolor{none_1}{HTML}{079FAB}        
\definecolor{none_2}{HTML}{E0BA32}        
\definecolor{guide_1}{HTML}{545E6F}       
\definecolor{guide_2}{HTML}{DD996F}       
\definecolor{hard_1}{HTML}{2F81DC}        
\definecolor{hard_2}{HTML}{26AC69}        
\definecolor{hard_3}{HTML}{8A3AA0}        
\colorlet{cons_color}{propose_color}
\colorlet{score_color}{hard_1}
\colorlet{b6_color}{none_1}\colorlet{b3_color}{none_2}
\colorlet{b7_color}{guide_1}\colorlet{b5_color}{guide_2}
\colorlet{b1_color}{hard_1}\colorlet{b2_color}{hard_2}\colorlet{b4_color}{hard_3}
\definecolor{MedBlue}{HTML}{3594cc}\definecolor{MedOrange}{HTML}{f0b077}

\definecolor{fig_plan}{HTML}{0072B2}     
\definecolor{fig_nominal}{HTML}{8A9198}  
\definecolor{fig_hazard}{HTML}{C44E52}   
\definecolor{fig_hazfill}{HTML}{F4E0E1}
\definecolor{fig_wall}{HTML}{EDF0F3}
\definecolor{fig_walledge}{HTML}{CAD0D7}
\definecolor{fig_safefill}{HTML}{F2F6F9}

\DeclareFontFamily{OT1}{figMono}{\hyphenchar\font=-1}
\DeclareFontShape{OT1}{figMono}{m}{n}{<-> fonts/figmono}{}
\DeclareFontShape{OT1}{figMono}{bx}{n}{<-> fonts/figmonob}{}
\DeclareFontShape{OT1}{figMono}{b}{n}{<-> fonts/figmonob}{}
\DeclareFontFamily{TS1}{figMono}{}
\DeclareFontShape{TS1}{figMono}{m}{n}{<-> fonts/figmonots}{}
\DeclareFontShape{TS1}{figMono}{bx}{n}{<-> fonts/figmonots}{}
\DeclareFontShape{TS1}{figMono}{b}{n}{<-> fonts/figmonots}{}
\pdfmapline{=figmono plxMono " AutoEnc_lxmhqhn4noowsq3ysicmnkmtrh ReEncodeFont " <[fonts/figmono.enc <fonts/figmono.pfb}
\pdfmapline{=figmonob plxMono-Bold " AutoEnc_lxmhqhn4noowsq3ysicmnkmtrh ReEncodeFont " <[fonts/figmono.enc <fonts/figmonob.pfb}
\pdfmapline{=figmonots plxMono " AutoEnc_koxhodrjcejglmtd3fkkpro4md ReEncodeFont " <[fonts/figmonots.enc <fonts/figmono.pfb}
\newcommand{\figfont}[1]{\fontencoding{OT1}\fontfamily{figMono}\fontseries{m}%
  \fontshape{n}\fontsize{#1}{#1}\selectfont}
\newcommand{\figfontb}[1]{\fontencoding{OT1}\fontfamily{figMono}\fontseries{b}%
  \fontshape{n}\fontsize{#1}{#1}\selectfont}

\newcommand{\figup}{{\fontencoding{TS1}\fontfamily{figMono}\selectfont\textuparrow}}
\newcommand{\figdown}{{\fontencoding{TS1}\fontfamily{figMono}\selectfont\textdownarrow}}

\pgfplotsset{
  asap/.style={
    scale only axis, clip=false,
    axis x line*=bottom, axis y line*=left,
    axis line style={line width=0.6pt, draw=fig_ink!85},
    ymajorgrids, xmajorgrids=false,
    grid style={fig_grid, line width=0.5pt},
    tick align=outside, major tick length=2.2pt,
    tick style={line width=0.6pt, fig_ink!85},
    ticklabel style={font=\figfont{8}, text=fig_ink},
    xticklabel={\pgfmathprintnumber[assume math mode=true]{\tick}},
    yticklabel={\pgfmathprintnumber[assume math mode=true]{\tick}},
    label style={font=\figfont{9}, text=fig_ink},
    title style={font=\figfont{9}, text=fig_ink},
    legend style={draw=none, fill=none, font=\figfont{8}, text=fig_ink,
                  /tikz/every even column/.append style={column sep=0.30em}},
    legend image code/.code={\draw[mark repeat=2, mark phase=2, #1]
      plot coordinates {(0cm,0cm) (0.15cm,0cm) (0.30cm,0cm)};},
    legend image post style={mark size=2.4pt, line width=1.1pt},
  },
  asap line/.style 2 args={color=#1, line width=0.9pt, mark=#2, mark size=1.8pt,
    mark options={solid, fill=#1, draw=fig_edge, line width=0.4pt}},
  asap line none/.style 2 args={asap line={#1}{#2},
    dash pattern=on 2.4pt off 1.7pt},
  asap ours/.style={color=propose_color, line width=1.4pt, mark=*, mark size=2.2pt,
    mark options={solid, fill=propose_color, draw=fig_edge, line width=0.45pt}},
}

\newsavebox{\asaprowbox}
\newlength{\asaprowh}

\newlength{\asapmazeh}

\newlength{\asaptabrowh}

%% file: sections/introduction.tex
\section{Introduction}
Generative models, such as diffusion models \citep{ho2020denoising} and flow matching \citep{lipman2022flow}, have recently emerged as a powerful paradigm for robotic trajectory planning. By learning trajectory distributions directly from data, they can naturally capture complex and multimodal behavior \citep{chi2025diffusion} in long-horizon tasks \citep{janner2022planning}. Unlike conventional trajectory-optimization planners \citep{kalakrishnan2011stomp, posa2014direct}, which rely on iterative online optimization and accurately learned system models, generative approaches \citep{janner2022planning, zheng2023guided} implicitly encode system dynamics and directly generate full trajectories at inference time. This enables scalable modeling of high-dimensional behaviors, supports task compositionality, and mitigates error accumulation from recursive rollout. 

Despite their promise, existing generative planning approaches typically suffer from one or both of two key limitations: slow multi-step inference and a lack of formal safety guarantees, as summarized in \Cref{tab:compare_related_work}. Rejection sampling discards unsafe trajectories but becomes ineffective when safe samples are rare under unseen constraints. Post-processing methods~\citep{maze2023diffusion} correct trajectories after generation but can substantially deviate from the learned distribution~\citep{romer2024diffusion}. Guidance-based approaches~\citep{kondo2024cgd, CG, mizuta2024cobl} steer samples toward safe regions, yet soft guidance may leave residual violations without formal guarantees. To obtain certified safety, recent methods regulate the generative dynamics of diffusion models \citep{xiao2023safediffuser} or flow models \citep{huang2026sad, yang2026safeflowmatcher} using control-theoretic mechanisms. However, these guarantees require repeated safety enforcement throughout multi-step generation, introducing substantial sampling overhead and limiting real-time planning.\looseness=-1 

To accelerate planning, recent works exploit the self-consistency property of generative models \citep{frans2025one, song2023consistency} to enable arbitrary-step sampling, including one-step generation. MeanFlow \citep{geng2025mean} further learns the average velocity over arbitrary time intervals, enabling efficient trajectory generation in robot manipulation \citep{sheng2026mp1} and autonomous driving \citep{wang2026meanfuser}. However, these methods lack formal safety guarantees. The safety mechanisms above are derived in continuous time~\citep{xiao2023safediffuser, huang2026sad}, so their guarantees do not directly extend to finite-step generation, particularly under few-step sampling with larger discretization errors. Consequently, guaranteeing safety under arbitrary-step generation remains an open problem, limiting its potential for real-time safe planning.

\input{tables/intro_method_compare.tex}

To address these limitations, we propose \textbf{A}rbitrary-Step \textbf{SA}fe \textbf{P}lanning (ASAP), a control-augmented MeanFlow framework for safe trajectory generation with any number of sampling steps. Guaranteeing safety for arbitrary-step generation is non-trivial because each MeanFlow update may span a large interval, so safety at the current sample does not generally imply safety at the next. ASAP directly constrains the next generated sample using an arbitrary-step affine barrier that approximates this discrete safety condition. Exploiting signed-distance geometry with a minimum-norm correction makes this approximation accurate, yielding zero conservatism. The resulting control is computed through a lightweight quadratic program, enabling safe planning with only a few, or even one, model inference and control solve, without retraining. For one-step generation, ASAP reduces to final-only SDF projection; for few-step generation, it goes beyond terminal projection by leveraging relaxed intermediate guidance to achieve superior task performance. As summarized in \Cref{tab:compare_related_work}, to our knowledge, ASAP is the first generative planner with formal safety guarantees for arbitrary-step generation, including the one-step case. We evaluate ASAP across navigation, locomotion, and manipulation benchmarks, where it satisfies all safety constraints while planning at least $20\times$ faster than safety-guaranteed baselines and over $200\times$ faster on a real-world robot-handing task, requiring only $0.04$\,s per plan. To evaluate scalability, we further test on the high-dimensional Adroit task, ASAP remains safe while planning over $2\times$ faster than safety-guaranteed baselines.\looseness=-1

%% file: tables/intro_method_compare.tex
\begin{wraptable}{r}{0.33\columnwidth}
\vspace{-0.9em}
\centering
\caption{\textbf{Methods overview.} Only ASAP ensures safety, arbitrary-step generation, and a formal guarantee.}
\label{tab:compare_related_work}
\vspace{0.2em}
\setlength{\tabcolsep}{2pt}
\renewcommand{\arraystretch}{0.96}
\footnotesize

\resizebox{\linewidth}{!}{%
\begin{tabular}{
    >{\centering\arraybackslash}m{2.7cm}
    >{\centering\arraybackslash}m{1.0cm} 
    >{\centering\arraybackslash}m{1.1cm} 
    >{\centering\arraybackslash}m{1.3cm} 
}
\toprule

\textbf{Generative Planners}
&
\shortstack[c]{\strut Safety\\constraint\strut}
&
\shortstack[c]{\strut Arbitrary\\step\strut}
&
\shortstack[c]{\strut Formal\\guarantee\strut} \\

\midrule
\shortstack[c]{SafeFlowMatcher\\ \citep{yang2026safeflowmatcher}}
& \cmark & \xmark & \cmark \\
\cmidrule{1-4}

\shortstack[c]{MeanFlow\\ \citep{geng2025mean}}
& \xmark & \cmark & \xmark \\
\cmidrule{1-4}

\shortstack[c]{CoBL-diffusion\\ \citep{mizuta2024cobl}}
& \cmark  & \xmark & \xmark \\
\cmidrule{1-4}

ASAP (Ours)
& \cmark & \cmark & \cmark \\

\bottomrule
\end{tabular}
}
\vspace{-0.2em}
\end{wraptable}

%% file: sections/related_work.tex
\section{Related Work}
\subsection{Planning with Diffusion-based and flow-based Generative Models}

Diffusion models~\citep{ho2020denoising, song2020score} and flow-based models~\citep{lipman2022flow, albergo2022building} have been increasingly adopted for data-driven robotic planning to learn complex expert behaviors from offline datasets~\citep{chi2025diffusion, huang2024toward}. Conditioned on the current state, these models can generate state-action trajectories directly~\citep{janner2022planning, zheng2023guided} or high-level trajectories tracked by downstream controllers~\citep{chi2025diffusion, ajay2022conditional}. However, conventional generative planners require iterative multi-step sampling, limiting their computational efficiency for real-time planning. Recent arbitrary-step generative models~\citep{frans2025one, geng2025mean} enable few- or one-step generation and have been applied to robotic manipulation and autonomous driving~\citep{sheng2026mp1, wang2026meanfuser}. Nevertheless, these fast generative planners have no mechanism to guarantee that the generated trajectories satisfy safety constraints, which is the gap addressed by ASAP.

\subsection{Constraint-Aware Planning with Generative Models}

To incorporate safety constraints into generative planning, existing methods broadly consider training-time and test-time mechanisms. Training-based approaches~\citep{ajay2022conditional, hung2025efficient} incorporate constraint information during learning to increase the likelihood of feasible trajectories. However, they require additional constraint-aware data and often generalize poorly to unseen constraints. Test-time approaches instead enforce constraints during or after generation. Post-processing methods~\citep{li2024diffusolve, maze2023diffusion} modify trajectories after sampling, but such corrections are unaware of the learned distribution and may cause substantial distributional deviation. More recent works introduce control-theoretic mechanisms directly into the generation. Constrained optimal-control layers are integrated into diffusion sampling~\citep{romer2024diffusion}, while control barrier functions~\citep{ames2016control} have been applied to diffusion-based planners~\citep{xiao2023safediffuser} and flow-based planners~\citep{yang2026safeflowmatcher, huang2026sad} to provide formal safety guarantees. However, these methods rely on repeated prediction and immediate safety correction throughout multi-step generation, incurring substantial sampling cost and limiting real-time applicability. In contrast, ASAP directly constrains each finite-step transition through a discrete-time barrier condition, providing safety guarantees for arbitrary-step generation, including a single step.

%% file: sections/problem_setting.tex
\section{Preliminaries and Problem Setting}
\subsection{Flow Matching}
Given a trajectory dataset $\mathcal{D}$ collected by an expert planner, continuous-time generative models learn to transport samples from a prior distribution $p_0$ to the expert trajectory distribution $p_1$. The transport is defined by a time-dependent vector field $\bm{v}_t(\bm{\tau}_t,t)$ through the ordinary differential equation (ODE)
\begin{equation} \label{eq:flow}
\dot{\bm{\tau}}_t
=
\bm{v}_t(\bm{\tau}_t,t),
\qquad
t\in[0,1],
\qquad
\bm{\tau}_0\sim p_0,
\end{equation}
where $\bm{\tau}_t$ denotes the nominal trajectory generated by the uncontrolled flow in \cref{eq:flow}. Flow matching \citep{lipman2022flow} can be applied to learn the vector field $\bm{v}_t(\bm{\tau}_t,t)$ with conditional flow matching loss $\mathcal{L}_{\text{CFM}}(\bm{\theta})=\mathbb{E}_{t,\bm{\tau}_0,\bm{\tau}_1}||\bm{v}^{\bm{\theta}}_t(\bm{I}_t,t)-\dot{\bm{I}}_t||^2_2$, where $\dot{\bm{I}}_t(\bm{\tau}_1,\bm{\tau}_0)\!=\!\dot{\alpha}(t)\bm{\tau}_1+\dot{\beta}(t)\bm{\tau}_0$ follows the scheduled interpolation $\bm{I}_t=\alpha(t)\bm{\tau}_1+\beta(t)\bm{\tau}_0$ with $\alpha,\beta$ such that $\alpha(0)\!=\!0, \alpha(1)\!=\!1$and $\alpha(t)\!+\!\beta(t)\!=\!1$. For trajectory planning, each sample $\bm{\tau}_1$ represents a complete planned trajectory. Thus, integrating \cref{eq:flow} with the learned vector field transports an initial sample $\bm{\tau}_0\sim p_0$ to a trajectory $\bm{\tau}_1\sim p_1$ that follows the expert distribution.
\subsection{Problem Statement}
Following the generative process in \cref{eq:flow}, $\bm{\tau}_1=\{\bm{\tau}_1(0),\ldots,\bm{\tau}_1(H-1)\}$ denotes the planned trajectory at the end of the process $t=1$, where $H\in\mathbb{N}$ is the planning horizon and $k=0,\ldots,H-1$ indexes trajectory states along this horizon, distinct from the generative time $t\in[0,1]$. For reliable real-world deployment, the planned trajectory along horizon $H$ must remain within a safe set $\mathbb{S}$, e.g., for collision avoidance, which may encode test-time constraints unseen in the training dataset $\mathcal{D}$.

\textbf{Objective.}
Given a pretrained generative planner $\bm{v}^{\bm{\theta}}_t$, our goal is to generate
\begin{equation} \label{eq:problem_setting}
\bm{\tau}_0 \sim p_0
\xrightarrow{\bm{v}^{\bm{\theta}}_t}
\bm{\tau}_1 \sim p_1,
\qquad
\bm{\tau}_1(k)\in\mathbb{S},
\quad
\forall k=0,\ldots,H-1.
\end{equation}
In particular, we require the planner to (i) generate trajectories efficiently with an arbitrary number of sampling steps, (ii) guarantee safety throughout the planned trajectory, and (iii) maintain competitive task performance under safety constraints.

%% file: sections/method.tex
\section{Method}\label{sec:method}
Existing safe flow planners establish safety by continuously enforcing control barrier function (CBF) conditions along the generative ODE. Such continuous-time guarantees do not directly extend to arbitrary-step sampling: each update spans a finite interval, so an instantaneous CBF condition may not ensure next-step barrier satisfaction. To address this gap, we propose ASAP, which starts from a discrete-time barrier formulation and derives an affine barrier condition guaranteeing safety for arbitrary sampling steps $N$, including $N=1$. In \Cref{sec:control_augmented_generative_model_planning}, we establish the control-augmented flow and its continuous-time safety guarantee, providing the foundation for arbitrary-step safety. In \Cref{sec:arb_safe_flow}, we derive an arbitrary-step affine barrier constraint from the discrete-time safety condition and establish formal safety guarantees, while \Cref{sec:analysis_of_conservatism} analyzes the approximation induced by the affine barrier and establishes zero conservatism for the SDF formulation.

\subsection{Control-Augmented Generative Model Planning}\label{sec:control_augmented_generative_model_planning}
To ensure satisfaction of safety for the generation process, we augment the nominal flow in \cref{eq:flow} at test time with a control input $\bm{u}_t$, yielding a controlled flow process  
\begin{equation} \label{eq:control_flow} 
     \dot{\bm{\tau}}_t=\frac{d}{dt}\bm{\tau}_t=\bm{v}_t(\bm{\tau}_t,t)+\bm{u}_t, 
    \quad 
    t \in [0,1], 
    \quad 
    \bm{\tau}_0\sim p_0,
\end{equation}
where $\bm{u}_t=0$ recovers the nominal flow generation process as in \cref{eq:flow}. The formulation in \cref{eq:control_flow} allows us to reinterpret the requirement in \cref{eq:problem_setting} as a control design problem: design $\bm{u}_t$ for test-time generation so that the planned trajectories satisfy the safety constraints.

\textbf{Control Barrier Constraints for Safety.} We formalize this control objective using control barrier functions (CBFs) \citep{ames2016control}, applied to each trajectory state $\bm{\tau}_t(k)$, $k=0,\ldots,H-1$ to enforce trajectory-level safety. Under the controlled dynamics in \cref{eq:control_flow}, the CBF condition prevents the barrier value from decreasing too rapidly near the safety boundary:
\begin{equation} \label{eq:cbf}
    h(\bm{\tau}_t(k)) + \dot{h}(\bm{\tau}_t(k)) \geq 0, 
    \quad
    \dot{h}(\bm{\tau}_t(k))\!=\!\nabla h(\bm{\tau}_t(k))^\top(\bm{v}_t(\bm{\tau}_t(k),t)\!+\!\bm{u}_t(k)),
\end{equation}
for all $k=0,\ldots,H-1$. Since the barrier constraint is enforced independently at each trajectory index $k$, we use $h(\bm{\tau}_t):=h(\bm{\tau}_t(k))$, and $\bm{u}_t:=\bm{u}_t(k)$ in the rest of the paper for simplicity.

\textbf{Time-Varying Relaxation of CBFs.} Although CBFs have been widely applied in dynamic systems, their safety condition can be relaxed for flow dynamics \cref{eq:control_flow} since safety is only required at the terminal time $t\!=\!1$. To avoid unnecessarily constraining intermediate generative states, we introduce a time-varying relaxed schedule $\rho_t=-\bar{h}(1-\alpha_t)$, where $\alpha_t\!:\![0,1] \!\rightarrow\! [0,1]$ is nondecreasing with $\alpha_0\!=\!0, \alpha_1\!=\!1$, and $\bar{h}\!\in\! \mathbb{R}_+$ is chosen such that
$\bar{h}\geq\max\{-h(\bm{\tau}_0),0\}$. We formulate the minimum-norm optimal control problem with the time-varying CBF condition to solve for $\bm{u}^*_t$
\begin{equation} \label{eq:QP_continuous_CBF}
\bm{u}^*_t
\coloneqq
\argmin_{\bm{u}_t\in\mathcal{F}_t}
\|\bm{u}_t\|^2,
\qquad
\mathcal{F}_t
\coloneqq
\left\{
\bm{u}_t
\,\middle|\,
h(\bm{\tau}_t)
+
\dot{h}(\bm{\tau}_t)
\ge
\rho_t+\dot{\rho}_t 
\right\}\text{ for all }
 t\in[0,1], 
\end{equation}
while minimally perturbing the original vector field $\bm{v}_t(\cdot,\cdot)$. The condition $\mathcal{F}_t$ in \cref{eq:QP_continuous_CBF} is most relaxed at $t\!=\!0$ and tightens to the safety requirement at $t\!=\!1$. This gives safety guarantees. \footnote{For a proof of \Cref{lemma:PTCBF-Flow} see Appendix \ref{append:proof}}

\begin{lemma}[\textbf{Continuous-time Safe Flow}] \label{lemma:PTCBF-Flow} 
If the CBF $h$ is differentiable at $\bm{\tau}_t$, applying the control $\bm{u}_t$ in \cref{eq:QP_continuous_CBF} to \cref{eq:control_flow} guarantees safety at $t=1$ for every initial condition $\bm{\tau}_0$.
\end{lemma}
\Cref{lemma:PTCBF-Flow} establishes safe generation for our control-augmented flow with the time-varying relaxation $\rho_t$. Extending this guarantee to arbitrary-step generation is non-trivial: safety relies on immediate correction, whereas large sampling intervals delay the next intervention. Thus, a continuous-time CBF condition may not ensure next-step safety if the step size is large. This motivates the discrete-time reformulation in \Cref{sec:arb_safe_flow}, which directly constrains the next generated sample.\looseness=-1

\input{figures/fig_overview.tex}

\subsection{Arbitrary-Step Flow with Safety}\label{sec:arb_safe_flow} 
To enable generation with arbitrary steps, we first consider the uncontrolled flow in \cref{eq:flow} and adopt MeanFlow \citep{geng2025mean} to learn the nominal average velocity $\bm{\mu}$ over a finite interval $[r,\!t]$ by
$
\bm{\mu}(\bm{\tau}_t,r,t)
\!\triangleq\!
\frac{1}{t-r}
\int_r^t
\bm{v}(\bm{\tau}_z,z)dz.
$
Differentiating it yields the MeanFlow identity
$\bm{\mu}(\bm{\tau}_{r},r,t) \!= \! \bm{v}(\bm{\tau}_{r},t)\! +\! (t-r)(\bm{v}(\bm{\tau}_{r},r)\partial_{\bm{\tau}}\bm{\mu}\!+\!\partial_{r}\bm{\mu})$, which allows us to learn the interval-averaged velocity $\!\bm{\mu}(\bm{\tau}_r,r,t)$ that summarizes the flow over $[r,t]$, replacing the repeated instantaneous-velocity $\bm{v}(\bm{\tau}_{r},t)$ predictions of standard flow matching in \cref{eq:flow}:\looseness=-1
\begin{align} \label{eq:meanflow_loss}
    &\mathcal{L}(\bm{\theta})=\mathbb{E}||\bm{\mu}^{\bm{\theta}}(\bm{\tau}_{r},r,t)-\text{sg}(\bm{\mu}_{\text{tgt}})||_2^2
    &\text{s.t.} \quad \ \bm{\mu}_{\text{tgt}}=\bm{v}+(t-r)(\bm{v}\partial_{\bm{\tau}}\bm{\mu}^{\bm{\theta}}+\partial_{r}\bm{\mu}^{\bm{\theta}}),
\end{align}
where the $\text{sg}$ is the stop-gradient operation.
By predicting the nominal average velocity over arbitrary intervals, MeanFlow enables few-step and even one-step trajectory generation via:
\begin{equation}\label{eq:mf_update}
    \bm{\tau}_{t}=\bm{\tau}_r+(t-r)\bm{\mu}^{\bm{\theta}}(\bm{\tau}_r,r,t).
\end{equation}

\textbf{Controlled MeanFlow for Arbitrary-Step Generation.} To enforce safety during arbitrary-step generation, following \cref{eq:control_flow}, we augment MeanFlow update~\cref{eq:mf_update} with a control input $\bm{u}_{t_i}$ that corrects the nominal sample toward constraint satisfaction. An $N$-step MeanFlow generation partitions $[0,1]$ into $N$ intervals, with $t_i\!=\!\frac{i}{N}$, $i\!=\!0,\ldots,N$, and sampling period $\Delta t\!=\!\frac{1}{N}$. For the $i$-th interval $[t_i,t_{i+1}]$, we denote the MeanFlow prediction by $\bm{\mu}^{\bm{\theta}}_{t_i}$ and formulate the control-augmented update as\looseness=-1
\begin{align}
    \text{\emph{Nominal Update:} } &\hat{\bm{\tau}}_{t_i}=\bm{\tau}_{t_i}+\bm{\mu}^{\bm{\theta}}_{t_i}\Delta t \label{eq:update_nominal} \\
    \hspace*{-0.8em}\text{\emph{Controlled Update:} }&\bm{\tau}_{t_{i+1}}=\hat{\bm{\tau}}_{t_i}+\bm{u}_{t_i}\Delta t,
    \quad i=0,\dots,N-1,\label{eq:update_control}
\end{align}
where \cref{eq:update_nominal} gives the nominal MeanFlow prediction, while $\bm{u}_{t_i}$ corrects it toward constraint satisfaction. To progressively tighten the safety requirement across discrete generation steps, the relaxation schedule $\rho_t$ defined in \Cref{sec:control_augmented_generative_model_planning} is evaluated at each sampling time: 
\begin{equation} \label{eq:discrete_time_varying_relax_term}
    \rho_{t_i}=-\bar{h}(1-\alpha_{t_i}),\quad -\bar{h}=\rho_{t_{0}=0}\leq\rho_{t_1}\leq\cdots\leq\rho_{t_N=1}=0, \quad \bar{h}\geq\max\{-h(\bm{\tau}_{t_0}),0\}
\end{equation}
\textbf{Arbitrary-Step Affine Barrier Constraints.} 
For the discrete updates in \cref{eq:update_nominal} and \cref{eq:update_control}, safety requires the next generated sample to satisfy 
$h(\bm{\tau}_{t_{i+1}})\ge\rho_{t_{i+1}}$. Since a discrete update may span a large interval $\Delta t$, the continuous-time CBF condition enforced at $t_i$ does not generally guarantee barrier satisfaction at $t_{i+1}$. Therefore, motivated by a discrete-time barrier condition~\citep{agrawal2017discrete}, we directly constrain the next-step generated sample $\bm{\tau}_{t_{i+1}}$ from \cref{eq:update_control} as
\begin{equation}\label{eq:discrete_cbf}
    h(\bm{\tau}_{t_{i+1}})\!=\!
h(\hat{\bm{\tau}}_{t_i}\!+\!\bm{u}_{t_i}\Delta t)
\!\geq\! \rho_{t_{i+1}}.
\end{equation}
Since the discrete barrier condition~\cref{eq:discrete_cbf} is generally nonlinear in $\bm{u}_{t_i}$,  we linearize it at the nominal MeanFlow update $\bm{u}_{t_i}\!=\!0$ and later show that, under our formulation, the linearization introduces no approximation error. This yields our arbitrary-step affine barrier constraint:\looseness=-1
\begin{equation} \label{eq:arb_Step_barrier_function}
    h(\hat{\bm{\tau}}_{t_i})+\nabla h(\hat{\bm{\tau}}_{t_i})^\top(\bm{u}_{t_i}\Delta t) \ge \rho_{t_{i+1}}.
\end{equation}
With the next-step threshold $\rho_{t_{i+1}}$ in \cref{eq:discrete_time_varying_relax_term}, we choose the $\bm{u}_{t_i}$ by the minimum-norm control as:
\begin{equation} \label{eq:QP_arb_Step_barrier_function}
    \bm{u}_{t_i} = \argmin_{\bm{u}} ||\bm{u}||^2 
    \quad
    \text{s.t. \cref{eq:arb_Step_barrier_function}}
\end{equation}
Because the barrier constraint is affine in $\bm{u}_{t_i}$, \cref{eq:QP_arb_Step_barrier_function} is a quadratic program (QP) that can be solved efficiently. We construct the CBF as a signed distance function (SDF) \citep{long2021learning, park2019deepsdf}, whose value gives the signed distance to the boundary of $\mathbb{S}$. \footnote{For a formal definition and more details of the SDF $h(\bm{\tau}_{t_i})$, see Appendix \ref{append:SDF}}. The key insight of ASAP is to exploit a geometric property of the SDF: at a differentiable point outside $\mathbb{S}$, $||\nabla h||=1$, $\nabla h$ points toward the nearest safe boundary, and $h$ varies linearly along this direction. When the barrier constraint is active, the minimum-norm solution of \cref{eq:QP_arb_Step_barrier_function} is aligned with $\nabla h$, as any orthogonal component increases the control norm without affecting the constraint. Consequently, the barrier value in \cref{eq:arb_Step_barrier_function} exactly matches the true next-step value $h(\bm{\tau}_{t_{i+1}})$. Hence, satisfying the affine constraint in~\cref{eq:arb_Step_barrier_function} guarantees $h(\bm{\tau}_{t_{i+1}})\geq\rho_{t_{i+1}}$, yielding the following results
\begin{theorem}[\textbf{Arbitrary-Step Safe Flow}] \label{th:arb_Step_safe_flow}
Given an initial condition $\bm{\tau}_{t_0}$ and if the SDF is differentiable for all $\hat{\bm{\tau}}_{t_i}$, the control input $\bm{u}_{t_i}$ in \cref{eq:QP_arb_Step_barrier_function} ensures that the solution $\bm{\tau}_{t_N}$ of \cref{eq:update_control} is safe.\looseness=-1\footnote{The proof of \Cref{th:arb_Step_safe_flow}, \Cref{corollary:one_step_safe_flow}, and \Cref{prop:robust_safe_flow} are provided in Appendix \ref{append:proof}}
\end{theorem}
Since \Cref{th:arb_Step_safe_flow} holds for arbitrary $N$, the one-step safety guarantee follows immediately.
\begin{corollary}[\textbf{One-Step Safe Flow}] \label{corollary:one_step_safe_flow}
Under the conditions of \Cref{th:arb_Step_safe_flow}, for $N=1$ and $\Delta t=1$, the control input $\bm{u}_{t_0}$ in \cref{eq:QP_arb_Step_barrier_function} ensures that the one-step solution $\bm{\tau}_{t_1}$ of \cref{eq:update_control} is safe.
\end{corollary}
\textbf{Robust Arbitrary-Step Safety.}
To account for practical inaccuracies, e.g., SDF approximation errors,
we extend \cref{eq:arb_Step_barrier_function} with a robustness margin
$r_{t_i}=t_i\bar e$, where $\bar e\geq0$ bounds the integrated barrier error.
The robust affine constraint becomes
\begin{equation}\label{eq:robust_arb_barrier}
    \hat h(\hat{\bm{\tau}}_{t_i})
    +\hat{\bm{g}}(\hat{\bm{\tau}}_{t_i})^\top
    (\bm{u}_{t_i}\Delta t)
    \geq \rho_{t_{i+1}}+r_{t_{i+1}},
\end{equation}
where $\hat h$ and $\hat{\bm{g}}$ denote the estimated SDF and gradient,
respectively.
\begin{proposition}[\textbf{Robust Arbitrary-Step Safety}]
\label{prop:robust_safe_flow}
If the integrated barrier error is bounded by $\bar e$,
the control input $\bm{u}_{t_i}$ solved by QP subject to
\cref{eq:robust_arb_barrier} ensures that the solution $\bm{\tau}_{t_N}$ of \cref{eq:update_control} is safe.
\end{proposition}

Our ASAP procedure is summarized in \Cref{alg:plan_alg}, and
\Cref{fig:overview} shows one sampling step. With setting $N=1$, planning requires only a single MeanFlow inference and a single QP solve, enabling real-time safe trajectory generation. Importantly, using an SDF introduces zero conservatism in the affine safety condition for any sampling step size, as analyzed in the next section.
\begin{remark}
The differentiability assumption is mild since SDFs are differentiable almost everywhere for sets with sufficiently regular boundaries \citep{gilbarg1977elliptic}. At nondifferentiable SDF points, generalized gradients~\citep{long2026sensor, glotfelter2017nonsmooth} can replace ordinary gradients by selecting a valid direction toward a nearest boundary point for the safety correction. Thus, this assumption can be handled with the practical use of SDF-based constraints \citep{long2021learning}.
\end{remark}
\begin{remark}
Unlike standard CBF control problems, the control input $\bm{u}_{t_i}$ in the flow dynamics \cref{eq:update_control} is unconstrained and acts directly on the generated trajectory. Consequently, each affine barrier constraint in \cref{eq:arb_Step_barrier_function} is always feasible whenever the SDF is differentiable. If $h(\hat{\bm{\tau}}_{t_i})\geq\rho_{t_{i+1}}$, $\bm{u}_{t_i}\!=\!\bm{0}$ is feasible; otherwise, since a differentiable SDF satisfies $\|\nabla h(\hat{\bm{\tau}}_{t_i})\|\!>\!0$, a correction along $\nabla h(\hat{\bm{\tau}}_{t_i})$ can always satisfy the constraint with equality. Thus, feasibility follows directly from the SDF geometry and the integrator-like generation dynamics; see Appendix~\ref{append:proof_feasibility} for the formal result.\looseness=-1
\end{remark}

\subsection{Conservatism Analysis}\label{sec:analysis_of_conservatism}
The previous section enforces arbitrary-step safety using the affine barrier condition in \cref{eq:arb_Step_barrier_function}. Here, conservatism refers to the mismatch between the barrier value used by this condition and the true next-step barrier value, which may induce unnecessary safety correction, rather than distributional deviation. For the exact SDF used by ASAP, the minimum-norm correction follows the SDF gradient, along which the SDF varies linearly, eliminating this mismatch for any sampling step size. We further analyze general CBFs in Appendix~\ref{append:general_barrier}, where the approximation error is bounded.

\textbf{Safety Conservatism.} With the affine barrier function in \cref{eq:arb_Step_barrier_function}, the value at next step is approximated by $\hat{h}(\hat{\bm{\tau}}_{t_i})\!=\!h(\hat{\bm{\tau}}_{t_i})\!+\! \nabla h(\hat{\bm{\tau}}_{t_i})^\top(\bm{u}_{t_i}\Delta t)$, and the resulting $\bm{u}_{t_i}$ obtained by \cref{eq:QP_arb_Step_barrier_function} updates $\hat{\bm{\tau}}_{t_i}\rightarrow\bm{\tau}_{t_{i+1}}$. We define the total conservatism $\bar{C}$ as the constraint residual summed over the $N$ steps 
\begin{equation} \label{eq:conservatism_term}
    \bar{C}=\sum_{i=0}^{N-1}C_i, \quad C_i = h(\bm{\tau}_{t_{i+1}}) - \hat{h}({\hat{\bm{\tau}}}_{t_i}),
\end{equation}
Based on the properties of SDF geometry described in \Cref{sec:arb_safe_flow}, we can obtain the following result\looseness=-1
\begin{theorem}[\textbf{Zero SDF Conservatism}] \label{th:zero_conservatism}
If the SDF $h$ is differentiable at $\hat{\bm{\tau}}_{t_i}$ for all $i$, then the control input $\bm{u}_{t_i}$ from \cref{eq:QP_arb_Step_barrier_function} yields $\bar{C}=0$ for any number of sampling steps $N$. \footnote{The proof of \Cref{th:zero_conservatism} and the corresponding validation are provided in Appendix \ref{append:proof_conservatism} and \ref{append:conservative}.\looseness=-1} 
\end{theorem}

If the affine barrier constraint is inactive, $\bm{u}_{t_i}=0$ and the approximation error is zero. If active, the minimum-norm correction follows the SDF gradient, along which the SDF varies linearly until the boundary. Therefore, the barrier condition remains exact over the finite-step correction, so the predicted and true next-step barrier values coincide, yielding zero conservatism for any $N$ step. This exact approximation avoids unnecessary safety corrections, helping preserve the original plan.

%% file: figures/fig_overview.tex
\newsavebox{\asapfigbox}
\newlength{\asapfigh}
\newlength{\asaptopw}\setlength{\asaptopw}{0.70in}
\newlength{\asapbotw}\setlength{\asapbotw}{1.30in}
\def\asapdiv{6.6pt}   
\newlength{\asapcapskip}\setlength{\asapcapskip}{7pt}
\newlength{\asaptoppad}\setlength{\asaptoppad}{3.2pt}
\newlength{\asapbotpad}\setlength{\asapbotpad}{-2.8pt}
\begin{figure}[t]
\centering
\begin{lrbox}{\asapfigbox}
\begin{minipage}[t]{0.56\textwidth}\vspace{0pt}%
\centering
\begin{tikzpicture}[
  x=1cm, y=1cm,
  panel/.style ={inner sep=0pt, outer sep=0pt, draw=fig_walledge,
                 line width=0.4pt},
  blab/.style  ={font=\fontsize{7.5}{8.5}\selectfont, align=center, inner sep=0pt, text depth=0pt},
  note/.style  ={font=\figfont{6.5}, text=fig_muted, align=center, inner sep=0pt},
  flow/.style  ={-{Stealth[length=4.0pt, width=3.2pt]}, line width=0.9pt,
                 fig_muted},
  qp/.style    ={-{Stealth[length=5.0pt, width=4.2pt]}, line width=1.4pt,
                 propose_color},
]

\node[panel] (p1) at (0,0) {\includegraphics[width=\asaptopw]{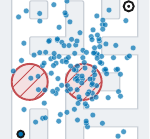}};
\node[panel, right=0.72cm of p1] (p2)
     {\includegraphics[width=\asaptopw]{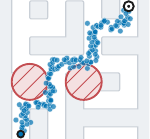}};
\node[panel, right=0.72cm of p2] (p3)
     {\includegraphics[width=\asaptopw]{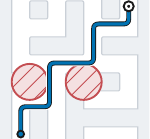}};

\draw[flow] ([xshift=1.0mm]p1.east) -- ([xshift=-1.0mm]p2.west);
\draw[flow] ([xshift=1.0mm]p2.east) -- ([xshift=-1.0mm]p3.west);
\node[note, above=0.8mm] at ($(p1.east)!0.5!(p2.west)$) {Step};
\node[note, above=0.8mm] at ($(p2.east)!0.5!(p3.west)$) {Step};

\node[font=\figfontb{7.5}, text=fig_ink, anchor=south]
  at ([yshift=0.8mm]$(p1.north)!0.5!(p3.north)$)
  {Generation in arbitrary steps\ {\scriptsize$(N\ge1)$}};

\node[blab, below=0.55mm of p1] {{\figfont{7.5}Noise}};
\node[blab, below=0.55mm of p2] (l2) {{\figfont{7.5}Intermediate}};
\node[blab, below=0.55mm of p3] {{\figfont{7.5}Safe plan}};

\coordinate (dv) at ($(p1.south west)!0.5!(p3.south east)$);
\coordinate (dv) at ([yshift=-\asapdiv]dv|-l2.south);
\draw[fig_walledge, line width=0.5pt] (p1.west|-dv) -- (p3.east|-dv);
\node[font=\figfontb{7.5}, text=fig_ink, fill=white, inner xsep=4pt,
      inner ysep=0.6pt] at (dv) {One step: propose, then correct};

\node[panel, anchor=north, yshift=-1.3mm] (q1)
  at ($(p1.west|-dv)!0.5!(p3.east|-dv)+(-2.02cm,0)$)
  {\includegraphics[width=\asapbotw]{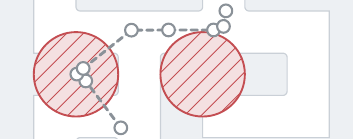}};
\node[panel, anchor=north, yshift=-1.3mm] (q2)
  at ($(p1.west|-dv)!0.5!(p3.east|-dv)+(2.02cm,0)$)
  {\includegraphics[width=\asapbotw]{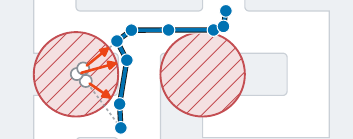}};

\draw[qp] ([xshift=0.8mm]q1.east) -- ([xshift=-0.8mm]q2.west);
\node[font=\figfontb{7}, text=propose_color, anchor=south, inner sep=1.0pt]
  at ($(q1.east)!0.5!(q2.west)$) {QP};

\node[blab, below=0.6mm of q1]
  {{\figfont{7.5}Proposal}\,{\scriptsize$\hat{\bm{\tau}}_{t_i}$}\\[1.2pt]
   {\figfont{6.5}\textcolor{fig_muted}{3 of 9 shown violate}}};
\node[blab, below=0.6mm of q2]
  {{\figfont{7.5}After the QP}\,{\scriptsize$\bm{\tau}_{t_{i+1}}$}\\[1.2pt]
   {\figfont{6.5}\textcolor{propose_color}{3 moved, 6 kept}}};
\end{tikzpicture}

\captionsetup{font=small,skip=\asapcapskip}
\vspace{0.3em}
\caption{\textbf{Schematic overview of ASAP.} \emph{Top:} $\bm{\tau}_{t_0}\!\sim\!p_0$
becomes a safe plan in any $N\!\ge\!1$ steps. \emph{Bottom:} the QP minimally
corrects MeanFlow's $H$-state proposal, restoring the barrier for constraints
unseen in training (red).}
\label{fig:overview}
\end{minipage}%
\end{lrbox}%
\setlength{\asapfigh}{\ht\asapfigbox}\addtolength{\asapfigh}{\dp\asapfigbox}%
\addtolength{\asapfigh}{\asapbotpad}%
\noindent\mbox{%
\usebox{\asapfigbox}%
\hspace{0.025\textwidth}%
\begin{minipage}[t][\asapfigh][s]{0.415\textwidth}\vspace{0pt}%
\input{figures/alg_asap.tex}%
\end{minipage}}
\end{figure}

%% file: figures/alg_asap.tex
\small\raggedright
\vspace*{\asaptoppad}
\hrule height 0.9pt
\vspace{3pt}
\begingroup
\captionsetup{type=algorithm, skip=3pt, justification=raggedright,
              singlelinecheck=false}
\caption{\textbf{Planning by ASAP.}}
\label{alg:plan_alg}
\endgroup
\vspace{1.5pt}\hrule height 0.4pt\vspace{3pt}
\begingroup
\renewcommand{\algorithmiccomment}[1]{\hfill{\footnotesize\textcolor{fig_muted}{#1}}}
\renewcommand{\algorithmicrequire}{\textbf{Input:}}
\renewcommand{\algorithmicensure}{\textbf{Output:}}
\noindent\parbox{\linewidth}{\raggedright\setlength{\baselineskip}{14.4pt}%
\begin{algorithmic}[1]
    \REQUIRE $\!\bm{\mu}^{\bm{\theta}}_{t_i}\!$, steps $N$, barrier $h$, $\rho_{t_i}$ by
             \cref{eq:discrete_time_varying_relax_term}\looseness=-1
    \STATE $\bm{\tau}_{t_0=0}\sim\mathcal{N}(\bm{0},\bm{I})$, $\Delta t=1/N$
    \FOR{$i=0,\cdots,N-1$}
        \STATE $\hat{\bm{\tau}}_{t_i}=\bm{\tau}_{t_i}+\bm{\mu}^{\bm{\theta}}_{t_i}\Delta t$
               \COMMENT{eq.\,(\ref{eq:update_nominal})}
        \STATE $\bm{u}_{t_i}\leftarrow\mathrm{QP}(\hat{\bm{\tau}}_{t_i},\rho_{t_{i+1}})$
               \COMMENT{eq.\,(\ref{eq:QP_arb_Step_barrier_function})}
        \STATE $\bm{\tau}_{t_{i+1}}=\hat{\bm{\tau}}_{t_i}+\bm{u}_{t_i}\Delta t$
               \COMMENT{eq.\,(\ref{eq:update_control})}
    \ENDFOR
    \ENSURE $\bm{\tau}_{t_N=1}$, with $\bm{\tau}_1(k)\in\mathbb{S}$ for all $k$
\end{algorithmic}}
\endgroup
\vspace{3pt plus 1fil}
\hrule height 0.4pt
\vspace{3pt}
\noindent\parbox{\linewidth}{\raggedright\footnotesize
\textbf{Cost per step:} one forward pass and one QP.\par}\par
\vspace{3pt}
\noindent\parbox{\linewidth}{\raggedright\footnotesize
\textbf{Guarantee:} \Cref{th:arb_Step_safe_flow} --- the returned
$\bm{\tau}_1$ satisfies every constraint, for every $N$.\par}\par
\vspace{1.2pt}
\hrule height 0.9pt

%% file: sections/experiments.tex
\section{Experiments} \label{sec:exp}
\subsection{Experiment Setup}
We evaluate ASAP on a variety of simulation benchmarks with test-time constraints as follows.
\begin{itemize}[left=0pt, itemsep=1pt, topsep=1pt, parsep=0pt, partopsep=0pt]
    \item \textbf{Maze2d} \citep{maze_env} considers point-mass navigation from an initial state to a target location. We introduce two previously unseen obstacles as state constraints while preserving at least one feasible route to the goal. The dataset is generated using a navigation planner in the large maze. \looseness=-1
    \item \textbf{Hopper} \citep{maze_env} is a locomotion benchmark in which an agent must move forward efficiently. We impose a torso-height constraint, which conflicts with forward locomotion. We use the Med-Expert (ME) dataset, combining expert and partially trained soft actor-critic trajectories.\looseness=-1
    \item \textbf{KUKA Block-Stacking} \citep{janner2022planning} is a 7-DoF robotic manipulator tasked with stacking multiple blocks. We introduce 10\% tighter joint-space constraints to prevent self-collision during manipulation. The dataset is generated with the PDDLStream planner \citep{garrett2020pddlstream}.
\end{itemize}
\textbf{Evaluation Metrics.} We quantify trajectory safety violation (s-vio) by the minimum constraint margin $-\min_k \min\{h_k(\bm{\tau}_1),0\},$
where $0$ indicates constraint satisfaction and larger values correspond to larger violations. Sampling time is measured as the total planning time (seconds), reflecting the capability for real-time deployment. Task performance is evaluated using normalized total rewards (score) for D4RL tasks, binary success rewards for block stacking following \cite{janner2022planning}. All results are evaluated on the same hardware, and the settings of baseline sampling steps follow \citep{xiao2023safediffuser} for a fair planning-time comparison. Further details are provided in Appendix~\ref{append:env}.\looseness=-1

\textbf{Baselines.} We compare our proposed ASAP with the following generative planning baselines:
\begin{itemize}[left=0pt, itemsep=1pt, topsep=1pt, parsep=0pt, partopsep=0pt]
    \item \textbf{MeanFlow (MF)} \citep{geng2025mean} generates trajectories with a pretrained MeanFlow. \looseness=-1
    \item \textbf{SafeFlowMatcher (SFM)} \citep{yang2026safeflowmatcher} generates a trajectory by a pretrained flow model in the prediction phase, and steers the trajectory toward a safe region by CBF in the correction phase. \looseness=-1
    \item \textbf{SAD-Flower (SAD-F)} \citep{huang2026sad} trains a flow model and generates trajectories with a control-augmented framework to incorporate test-time safety constraints. \looseness=-1
    \item \textbf{Diffuser} \citep{janner2022planning} is the pioneer work using a diffusion model to generate trajectories.
    \item \textbf{Classifier Guidance (CG)} \citep{CG} guides diffusion sampling using constraint gradients to steer generated trajectories toward safe regions during sampling.
    \item \textbf{SafeDiffuser (S-Diff.)}  \citep{xiao2023safediffuser} combines diffusion planning with CBF-based projection to enforce state constraints at test time throughout each denoise step.
    \item \textbf{CoBL} \citep{mizuta2024cobl} incorporates physical constraints into diffusion sampling through shaping control barrier functions and control Lyapunov function as guidance.
\end{itemize}

\input{tables/main_results}

\subsection{Safety and Efficiency Across Simulation Tasks}
For a fair evaluation, we compare all baselines under the same hardware and optimization software. As shown in \Cref{tab:main_result}, ASAP achieves the lowest computational cost among safety-guaranteed methods, planning at least $20\times$ faster while matching their task performance. MF is marginally faster with single-step sampling but provides no safety guarantees.
\looseness=-1

In the Maze2d, ASAP achieves competitive performance while satisfying all safety constraints, showing that safety can be enforced without sacrificing performance when the constraint is compatible with the task, e.g., obstacles do not fully block the path. Other control-theoretic baselines also ensure safety, yet have higher planning costs due to repeated corrections during multi-step generation.\looseness=-1 

In the Hopper task, ASAP guarantees safety with the lowest computational cost while maintaining competitive performance, despite the height constraint restricting low-posture motions for faster locomotion. Other control-theoretic baselines also ensure safety but require longer planning time, while SFM still violates the constraint due to its slack variable. These results further show that ASAP efficiently enforces hard safety constraints without severely degrading task performance.

Similar trends are observed in the KUKA block-stacking task, where all methods require more generation steps to achieve strong performance. ASAP also benefits from slightly more steps, but still requires far fewer corrections than competing safety-guaranteed methods, remaining over $250\times$ faster. This shows that ASAP can increase computation when beneficial for more complex tasks while retaining substantial efficiency. The further computational detail is reported in Appendix~\ref{append:comp_hor_con}.
\looseness=-1

\subsection{Real-World Robotic Planning}
We evaluate ASAP on a real-world service task where a Franka robot hands an ice cream to a customer through a shop window, as shown in \Cref{fig:real_world_setup}. A juicer is introduced as an obstacle in the test time to emulate constrained workspace geometry. We conservatively approximate the juicer with an enlarged ellipsoid to provide additional safety clearance. For hardware safety, the physical juicer is removed during potentially unsafe executions, while its geometry is retained as an SDF for collision evaluation. Demonstrations are collected by teleoperation, with the robot picking the ice cream from a plate and delivering it to the human hand. The generated trajectories are tracked using a low-level impedance controller. At deployment, we assume a perception pipeline, e.g., a motion-capture system, provides accurate estimates of the ice-cream and target-hand positions.

Safety is measured using the normalized signed distance $h_k=\operatorname{sdist}(\mathcal R_k,\mathcal O)/h_{\max}$, where $\mathcal R_k$ and $\mathcal O$ denote the robot and obstacle
geometries, respectively, approximated by ellipsoids. The safety violation is defined as
$s_{\mathrm{vio}}=-\min_k\min\{h_k,0\}$,
normalized to $[0,1]$.
 Task performance is based on final placement error, with a score of $1$ for delivery within $1.0$~cm of the target and higher scores for greater accuracy; the tolerance accounts for control error. We conduct $20$ trials across two initial plate locations and three target hand positions, which define the ice-cream delivery targets, with trials approximately evenly distributed across these configurations. We compare against MF, CG, and S-Diff., which perform best in planning time, task score, and safety, respectively, on a similar manipulation task for the KUKA block-stacking simulation.

As shown in \Cref{tab:real_world}, ASAP matches the best safety performance of S-Diff. and the best task score of MF with only $0.04$\,s. MF needs less time but cannot guarantee safety under the unseen obstacle, while S-Diff. ensures safety at higher computational cost. These results demonstrate that ASAP enables efficient safe planning under unseen real-world constraints, with details in Appendix~\ref{append:real_world}.\looseness=-1

\subsection{How ASAP Enables Fast and Safe Planning}
\textbf{Performance of ASAP with Different Sampling Steps.}
We evaluate ASAP and all baselines across sampling steps $N$ on the KUKA block-stacking task. As shown in \Cref{fig:nfe_score}, ASAP and MF first exceed a task score of $1$ at $N\!=\!3$ and achieve the best performance for $N\!\le\!10$. This demonstrates that arbitrary-step safe generation enables strong performance with few steps while preserving safety. In contrast, most baselines require around $10$ steps to exceed a score of $1$, with some improving further at higher computational cost. MF degrades at larger $N$, likely due to training--inference mismatch in the pretrained MeanFlow model, which is optimized primarily for few-step generation. ASAP inherits this behavior from its MeanFlow backbone while remaining safe at every $N$.\looseness=-1

\input{figures/fig_real_world.tex}
\input{figures/fig_steps_and_maze.tex}

\textbf{Handling Multiple Safety Constraints.}
We evaluate ASAP under multiple safety constraints by increasing the number of obstacles in Maze2d from two to five. Multiple obstacles can be represented by a single SDF over the union of unsafe sets, or equivalently by selecting the nearest active obstacle SDF at differentiable points, with details in Appendix~\ref{append:SDF}. As shown in \Cref{fig:multi_constraint}, ASAP generates collision-free trajectories satisfying all constraints while maintaining a similar planning time of approximately $0.02$\,s on average. This low computational overhead is enabled by the lightweight QP formulation, demonstrating the flexibility of ASAP in handling multiple safety constraints. 

\textbf{Affine Barrier vs. Direct Post-Optimization.}
We compare ASAP with a naive post-optimization baseline to isolate the computational benefit of our QP formulation. Both use the same MeanFlow model and non-QP-capable solver for fairness. Post-optimization directly optimizes the trajectory under the original safety constraint, yielding a quadratically constrained quadratic program in Maze2d, whereas ASAP linearizes the barrier into an affine constraint and a lightweight QP. As shown in \Cref{tab:qp_vs_post}, both achieve comparable task performance and satisfy the constraints, while ASAP roughly halves planning time, demonstrating the efficiency of the affine formulation.\looseness=-1

\input{tables/tables_456.tex}

\subsection{How Reliable Is ASAP}
\textbf{Handling Stricter Test-Time Constraints.}
We evaluate ASAP under increasingly strict unseen constraints in the KUKA task by progressively tightening the feasible joint range for self-collision avoidance. As shown in \Cref{fig:constraint_levels}, ASAP satisfies all constraint levels without retraining, at the lowest planning cost among methods that stay safe. In contrast, CoBL, Diffuser, CG, and MF exhibit increasing violations as the constraints become stricter. Among methods that remain safe, ASAP also achieves the highest task score. The score decreases at the $30\%$ constraint level because the tightened joint limits begin to restrict motions required for task completion.

\textbf{Sensitivity Analysis of Hyperparameters.}
We evaluate the sensitivity of ASAP to $\bar{h}$, which controls the initial safety relaxation in \cref{eq:discrete_time_varying_relax_term} with $5$ sampling steps, by varying it from the default $1.0$ to $3.0$ while respecting the theoretical minimum. The maximum initial constraint violation across all evaluated trajectory states is $-0.8751$, satisfying the theoretical condition $\bar{h}\geq\max_k\{-h(\bm{\tau}_{t_0}(k)),0\}$ for all tested values. As shown in \Cref{tab:abla_sensitivity}, ASAP remains safe across all tested values of $\bar{h}$ with similar task performance. These results show that ASAP is robust to $\bar{h}$ and does not require precise hyperparameter tuning.\looseness=-1

\textbf{Scalability to High-Dimensional Task.}
To evaluate scalability, we test ASAP on the D4RL Adroit Relocate task, a dexterous manipulation benchmark with 39-dimensional states and 30-dimensional actions. We impose collision-avoidance constraints based on the arm position. As shown in \Cref{tab:abla_highdim}, ASAP is the fastest safety-guaranteed method and satisfies all constraints. MF plans faster but violates safety, while other control-theoretic baselines also ensure safety at over $2\times$ higher computational cost. These results show that ASAP scales effectively to high-dimensional tasks while preserving safety and efficiency. Further details are provided in Appendix~\ref{append:append_adroit}.

\textbf{SDF Projection Strategies.}
To evaluate ASAP's SDF-based safety correction strategy, we compare it with MeanFlow using final-only SDF projection (Final-only) and full SDF projection at every step (Full). We focus on KUKA ($N\!=\!4$) and real-world handover ($N\!=\!2$), where multi-step generation allows us to evaluate the benefit of intermediate guidance. As shown in \Cref{tab:abla_cbf_projection}, all methods satisfy the safety constraints, while ASAP achieves the highest reported task scores. In particular, ASAP improves the KUKA score from $1.15$ (Final-only) and $1.45$ (Full) to $1.75$, and the real-world score from $0.80$ (Final-only) and $0.70$ (Full) to $1.00$. These results highlight the benefit of ASAP's flexible safety guidance for multi-step generation, while retaining efficient and safe one-step planning.\looseness=-1

\textbf{Robustness to Noise.}
To evaluate ASAP's robustness to practical inaccuracies, we introduce Gaussian noise with standard deviations $\sigma\in\{0.1,0.4,0.8\}$ in Maze2d and set $\bar e=1$ in the robust barrier formulation \cref{eq:robust_arb_barrier}. As shown in \Cref{tab:robust_CBF}, ASAP maintains zero safety violations across all noise levels, with similar task scores and planning times. These results demonstrate that ASAP maintains safety and task performance under the tested noise levels with minimal computational overhead.

\textbf{Arbitrary-Step Safety Corrections.}
To evaluate whether existing safe planners can benefit from MeanFlow generation, we compare them using single-step (1) and their default multi-step (M) correction settings. As shown in \Cref{tab:compare_single_multi_step}, all baselines exhibit safety violations with single-step correction, while their default multi-step settings maintain safety at higher computational costs. Notably, even when SFM uses a single MeanFlow prediction, multiple subsequent safety corrections are required to maintain safety in our evaluation. In contrast, ASAP achieves zero safety violations with one MeanFlow prediction and correction in $0.06$\,s, while maintaining competitive task performance. These results highlight the importance of ASAP's discrete-time safety formulation, enabling efficient few-step generation with formal safety guarantees rather than relying on repeated corrections. \looseness=-1

\input{tables/abla_cbf_projection}
\input{tables/tables_abla_robustCBF}

%% file: tables/main_results.tex
\begin{table}[!t]
\caption{\textbf{Main results.} Safety violation (s-vio; $0$ means every constraint is satisfied), planning time (s), and task score of every method shown as mean$\pm$std on the navigation, locomotion, and manipulation tasks. ASAP satisfies every constraint while planning at least $20\times$ faster than the safety-guaranteed baselines. 
ASAP uses a QP solver for flexibility in handling multiple constraints. Closed-form results are reported in Appendix~\ref{append:closed_form}.
}
\label{tab:main_result}
\vspace{0.3em}
\resizebox{\textwidth}{!}{
\begin{tabular}{
    >{\centering\arraybackslash}m{1.6cm} 
    >{\centering\arraybackslash}m{1.3cm} 
    >{\centering\arraybackslash}m{1.4cm} 
    >{\centering\arraybackslash}m{1.4cm} 
    >{\centering\arraybackslash}m{1.4cm} 
    >{\centering\arraybackslash}m{1.4cm} 
    >{\centering\arraybackslash}m{1.4cm} 
    >{\centering\arraybackslash}m{1.4cm} 
    >{\centering\arraybackslash}m{1.4cm} 
    >{\centering\arraybackslash}m{1.4cm} 
}
\toprule
\textbf{Task}  & \textbf{Metric} & \textbf{MF} & \textbf{SFM} & \textbf{SAD-F}  & \textbf{Diffuser} & \textbf{CG} & \textbf{S-Diff.} & \textbf{CoBL} & \textbf{ASAP} \\
\midrule
\multirow{4}{2cm}{\centering Maze2d (Large)} 
& s-vio ($\downarrow$)  &0.42$\pm$0.29 & \textbf{0.00$\pm$0.00} & \textbf{0.00$\pm$0.00} &0.22$\pm$0.28&0.69$\pm$0.20& \textbf{0.00}$\bm{\pm}$\textbf{0.00} & \textbf{0.00}$\bm{\pm}$\textbf{0.00}& \textbf{0.00}$\bm{\pm}$\textbf{0.00} \\
\cmidrule{2-10}
& time ($\downarrow$) & \textbf{0.01}$\bm{\pm}$\textbf{0.00} & 15.50$\pm$0.75 & 35.84$\pm$2.51 &1.64$\pm$0.03& 1.83$\pm$0.03& 17.08$\pm$2.29&2.26$\pm$0.03 & 0.06$\pm$0.01 \\
\cmidrule{2-10}
& score ($\uparrow$) &$\!\textbf{1.46}\!\bm{\pm}\!\textbf{0.47}\!$ &1.30$\pm$0.67 &1.44$\pm$0.28& 1.38$\pm$0.51& 0.42$\pm 0.29$&$\!\textbf{1.46}\!\bm{\pm}\!\textbf{0.23}\!$ & 0.79$\pm$0.71& $\!\textbf{1.46}\!\bm{\pm}\!\textbf{0.47}\!$\\
\midrule\midrule
\multirow{4}{2cm}{\centering Hopper \quad (ME)} 
& s-vio ($\downarrow$)& 0.01$\pm$0.01 & 0.04$\pm$0.02 & \textbf{0.00}$\bm{\pm}$\textbf{0.00}  &0.04$\pm$0.04& 0.14$\pm0.06$ & \textbf{0.00}$\bm{\pm}$\textbf{0.00} & 0.04$\pm$0.04& \textbf{0.00}$\bm{\pm}$\textbf{0.00}\\
\cmidrule{2-10}
& time ($\downarrow$)& \textbf{0.01}$\bm{\pm}$\textbf{0.00} & 1.13$\pm$0.01 & 1.66$\pm$0.03  &0.15$\pm$0.00& 0.15$\pm0.00$& 1.03$\pm$0.02 &0.16$\pm$0.00&  0.05$\pm$0.00\\
\cmidrule{2-10}
& score ($\uparrow$)& 0.41$\pm$0.14 & 0.39$\pm$0.09 &0.43$\pm$0.11& \textbf{0.46}$\bm{\pm}$\textbf{0.04} & 0.42$\pm$0.08& 0.44$\pm$0.07& \textbf{0.46}$\bm{\pm} \textbf{0.05}$& 0.41$\pm$0.14\\
\midrule\midrule
\multirow{4}{2cm}{\centering KUKA \\ (Stacking)} 
& s-vio ($\downarrow$)& 0.30$\pm$0.09 &\textbf{0.00}$\bm{\pm}$\textbf{0.00} & \textbf{0.00}$\bm{\pm}$\textbf{0.00} &0.31$\pm$0.12& 0.11$\pm0.02$ & \textbf{0.00}$\bm{\pm}$\textbf{0.00} & 0.31$\pm$0.02& \textbf{0.00}$\bm{\pm}$\textbf{0.00}\\
\cmidrule{2-10}
& time ($\downarrow$)& \textbf{0.03}$\bm{\pm}$\textbf{0.00} & 38.14$\pm$0.65 & 37.21$\pm$0.08  &21.25$\pm$0.19& 22.02$\pm0.05$& 68.75$\pm$1.65 &25.01$\pm$0.21&  0.10$\pm$0.00\\
\cmidrule{2-10}
& score ($\uparrow$)& 1.65$\pm$1.11 & 1.35$\pm$0.65 &1.45$\pm$0.67& 1.30$\pm$0.90 & \textbf{1.75}$\bm{\pm} \textbf{0.94}$& 1.60$\pm$0.86& 0.30$\pm$0.46& \textbf{1.75}$\bm{\pm} \textbf{0.83}$\\
\bottomrule\bottomrule
\end{tabular}
}
\vspace{-0.2em}
\end{table}

%% file: figures/fig_real_world.tex
\begin{figure}[t]
\centering
\noindent\mbox{%
\begin{minipage}[t]{0.52\textwidth}\vspace{0pt}%
    \centering
    \includegraphics[height=2.9cm,width=\linewidth,trim=0 45bp 0 150bp,clip]{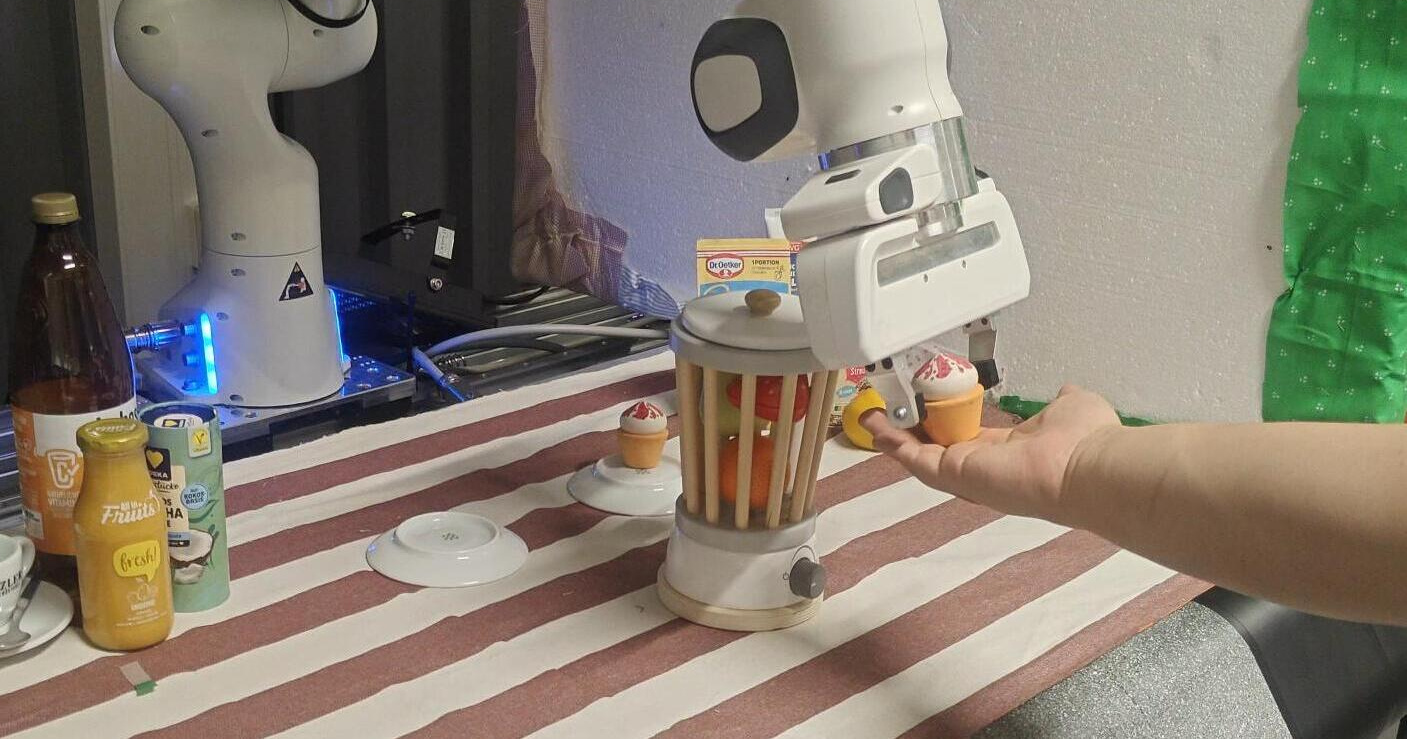}\par
    \vspace{3pt}
    \captionof{figure}{\textbf{Real-world handover.} A robot passes an ice cream past a juicer that is absent \mbox{from the demonstrations.}}
    \label{fig:real_world_setup}
\end{minipage}%
\hspace{0.02\textwidth}%
\begin{minipage}[t]{0.47\textwidth}\vspace*{-0.87pt}%
    \captionof{table}{\textbf{Real-world results.} ASAP ensures safety while planning over $200\times$ faster than the safety-guaranteed baseline. Bold marks the best value.}
    \label{tab:real_world}
    \vspace{1.5pt}
    \centering
    \input{tables/real_world_robot_table.tex}
\end{minipage}}
\end{figure}

%% file: tables/real_world_robot_table.tex
{\small
\setlength{\tabcolsep}{3pt}
\renewcommand{\arraystretch}{1.2425}
\begin{tabular*}{\linewidth}{@{\extracolsep{\fill}}lccc@{}}
\toprule
\textbf{Method} & \textbf{s-vio ($\downarrow$)} & \textbf{time ($\downarrow$)} & \textbf{score ($\uparrow$)} \\
\midrule
MF & 0.35$\pm$0.11 & \textbf{0.01$\pm$0.01} & 0.30$\pm$0.03 \\
CG & 0.27$\pm$0.12 & 2.18$\pm$0.23 & 0.99$\pm$0.01 \\
S-Diff. & \textbf{0.00$\pm$0.00} & 13.23$\pm$1.35 & 0.30$\pm$0.02 \\
\textbf{ASAP} & \textbf{0.00$\pm$0.00} & 0.04$\pm$0.01 & \textbf{1.00$\pm$0.01} \\
\bottomrule
\end{tabular*}}

%% file: figures/fig_steps_and_maze.tex
\usetikzlibrary{backgrounds}
\begin{figure}[t]
\centering
\def\asapH{44pt}%
\def\asapWa{102.5pt}
\def\asapTick{10.2pt}
\def\asapWc{50.5pt}%
\def\asapGab{28pt}
\def\asapGm{11pt}
\def\asapGcc{17pt}
\begin{tikzpicture}[
    ttl/.style={font=\figfont{7.5}, text=fig_ink, anchor=south, inner sep=0pt},
    xl/.style ={font=\figfont{7.5}, text=fig_ink, anchor=north, inner sep=0pt},
    plab/.style={font=\figfont{8.5}, text=fig_ink, anchor=base, inner sep=0pt},
    groupbox/.style={fill=white, draw=fig_rule!65, line width=0.4pt,
                     rounded corners=2pt},
]
\pgfplotsset{
    asaprow/.style={asap, height=\asapH, anchor=south west,
        ticklabel style={font=\figfont{6.5}},
        xticklabel style={font=\figfont{6.5}, anchor=base, yshift=-7.1pt}},
}
\begin{axis}[
    asaprow, name=A, at={(0pt,0pt)}, width=\asapWa,
    xmin=-0.3, xmax=7.3,
    xtick={0,1,2,3,4,5,6,7},
    xticklabels={1,2,3,4,5,10,10\textsuperscript{2},10\textsuperscript{3}},
    ymin=0, ymax=2, ytick={0,1,2},
    legend columns=8, legend to name=asapmethods,
    legend style={draw=none, fill=none, font=\figfont{7},
                  /tikz/row sep=-1pt,
                  /tikz/every odd column/.append style={column sep=0.15em},
                  /tikz/every even column/.append style={column sep=1.0em}},
]
    \draw[fig_rule!55, line width=0.45pt, dash pattern=on 0.6pt off 1.5pt]
        (axis cs:-0.3,1) -- (axis cs:7.3,1);
    \addplot[asap line none={none_1}{triangle*}]
        table[x expr=\coordindex, y=MF] {figures/data/abla_nfe_score.txt};
    \addlegendentry{MF}
    \addplot[asap line none={none_2}{square*}]
        table[x expr=\coordindex, y=Diffuser] {figures/data/abla_nfe_score.txt};
    \addlegendentry{Diffuser}
    \addplot[asap line={guide_1}{diamond*}]
        table[x expr=\coordindex, y=CG] {figures/data/abla_nfe_score.txt};
    \addlegendentry{CG}
    \addplot[asap line={guide_2}{pentagon*}]
        table[x expr=\coordindex, y=CoBL] {figures/data/abla_nfe_score.txt};
    \addlegendentry{CoBL}
    \addplot[asap line={hard_1}{triangle*}]
        table[x expr=\coordindex, y=SFM] {figures/data/abla_nfe_score.txt};
    \addlegendentry{SFM}
    \addplot[asap line={hard_2}{square*}]
        table[x expr=\coordindex, y=SADF] {figures/data/abla_nfe_score.txt};
    \addlegendentry{SAD-F}
    \addplot[asap line={hard_3}{diamond*}]
        table[x expr=\coordindex, y=S-Diff] {figures/data/abla_nfe_score.txt};
    \addlegendentry{S-Diff.}
    \addplot[asap ours]
        table[x expr=\coordindex, y=ASAP] {figures/data/abla_nfe_score.txt};
    \addlegendentry{ASAP}
\end{axis}
\node[ttl] (tA) at ([yshift=2pt]A.north) {Score (\figup)};
\node[xl]  (xA) at ([yshift=-13pt]A.south) {Sampling steps N};

\coordinate (bandN) at (tA.north);
\coordinate (bandS) at (xA.south);

\begin{axis}[
    asaprow, name=C1, at={($(A.east|-A.south)+(\asapGab,0)$)}, width=\asapWc,
    xmin=7, xmax=33, xtick={10,20,30}, xticklabels={10\%,20\%,30\%},
    ymin=0, ymax=1, ytick={0,0.5,1},
]
    \addplot[asap line none={none_1}{triangle*}, line width=2.4pt, mark size=2.8pt]
        table[x=constraint, y=MF] {figures/data/right_methods_safety_vs_constraint.txt};
    \addplot[asap line none={none_2}{square*}, line width=1.7pt, mark size=2.2pt]
        table[x=constraint, y=Diffuser] {figures/data/right_methods_safety_vs_constraint.txt};
    \addplot[asap line={guide_1}{diamond*}]
        table[x=constraint, y=CG] {figures/data/right_methods_safety_vs_constraint.txt};
    \addplot[asap line={guide_2}{pentagon*}, line width=1.1pt, mark size=1.6pt]
        table[x=constraint, y=CoBL] {figures/data/right_methods_safety_vs_constraint.txt};
    \addplot[asap line={hard_1}{triangle*}, line width=2.4pt, mark size=3.2pt]
        table[x=constraint, y=SFM] {figures/data/right_methods_safety_vs_constraint.txt};
    \addplot[asap line={hard_2}{square*}, line width=1.9pt, mark size=2.6pt]
        table[x=constraint, y=SADF] {figures/data/right_methods_safety_vs_constraint.txt};
    \addplot[asap line={hard_3}{diamond*}, line width=1.4pt, mark size=2.0pt]
        table[x=constraint, y=S-Diff] {figures/data/right_methods_safety_vs_constraint.txt};
    \addplot[asap ours, line width=1.0pt, mark size=1.4pt]
        table[x=constraint, y=ASAP] {figures/data/right_methods_safety_vs_constraint.txt};
\end{axis}
\begin{axis}[
    asaprow, name=C2, at={($(C1.east|-A.south)+(\asapGcc,0)$)}, width=\asapWc,
    xmin=7, xmax=33, xtick={10,20,30}, xticklabels={10\%,20\%,30\%},
    ymode=log, ymin=0.01, ymax=150, yminorticks=false,
    ytick={0.1,1,10,100}, yticklabels={0.1,1,10,100},
]
    \addplot[asap line none={none_1}{triangle*}] table[x=constraint, y=MF] {figures/data/right_methods_speed_vs_constraint.txt};
    \addplot[asap line none={none_2}{square*}, line width=2.0pt, mark size=2.5pt] table[x=constraint, y=Diffuser] {figures/data/right_methods_speed_vs_constraint.txt};
    \addplot[asap line={guide_1}{diamond*}, line width=1.4pt, mark size=1.9pt] table[x=constraint, y=CG] {figures/data/right_methods_speed_vs_constraint.txt};
    \addplot[asap line={guide_2}{pentagon*}, line width=0.9pt, mark size=1.4pt] table[x=constraint, y=CoBL] {figures/data/right_methods_speed_vs_constraint.txt};
    \addplot[asap line={hard_1}{triangle*}, line width=1.8pt, mark size=2.3pt] table[x=constraint, y=SFM] {figures/data/right_methods_speed_vs_constraint.txt};
    \addplot[asap line={hard_2}{square*}, line width=1.0pt, mark size=1.5pt] table[x=constraint, y=SADF] {figures/data/right_methods_speed_vs_constraint.txt};
    \addplot[asap line={hard_3}{diamond*}] table[x=constraint, y=S-Diff] {figures/data/right_methods_speed_vs_constraint.txt};
    \addplot[asap ours] table[x=constraint, y=ASAP] {figures/data/right_methods_speed_vs_constraint.txt};
\end{axis}
\begin{axis}[
    asaprow, name=C3, at={($(C2.east|-A.south)+(\asapGcc,0)$)}, width=\asapWc,
    xmin=7, xmax=33, xtick={10,20,30}, xticklabels={10\%,20\%,30\%},
    ymin=0, ymax=2, ytick={0,1,2},
]
    \addplot[asap line none={none_1}{triangle*}] table[x=constraint, y=MF] {figures/data/right_methods_score_vs_constraint.txt};
    \addplot[asap line none={none_2}{square*}] table[x=constraint, y=Diffuser] {figures/data/right_methods_score_vs_constraint.txt};
    \addplot[asap line={guide_1}{diamond*}] table[x=constraint, y=CG] {figures/data/right_methods_score_vs_constraint.txt};
    \addplot[asap line={guide_2}{pentagon*}] table[x=constraint, y=CoBL] {figures/data/right_methods_score_vs_constraint.txt};
    \addplot[asap line={hard_1}{triangle*}] table[x=constraint, y=SFM] {figures/data/right_methods_score_vs_constraint.txt};
    \addplot[asap line={hard_2}{square*}] table[x=constraint, y=SADF] {figures/data/right_methods_score_vs_constraint.txt};
    \addplot[asap line={hard_3}{diamond*}] table[x=constraint, y=S-Diff] {figures/data/right_methods_score_vs_constraint.txt};
    \addplot[asap ours] table[x=constraint, y=ASAP] {figures/data/right_methods_score_vs_constraint.txt};
\end{axis}
\node[ttl] at (tA.south-|C1.north) {s-vio (\figdown)};
\node[ttl] at (tA.south-|C2.north) {time (s, \figdown)};
\node[ttl] at (tA.south-|C3.north) {Score (\figup)};

\coordinate (groupNW) at ([xshift=-19pt,yshift=3pt]C1.west|-bandN);
\coordinate (groupSE) at ([xshift=3pt,yshift=-2pt]C3.east|-bandS);
\coordinate (groupCenter) at ($(groupNW)!0.5!(groupSE)$);
\node[xl] at (xA.north-|groupCenter) {Constraint level};

\path (A.north); \pgfgetlastxy{\asapbx}{\asapbN}%
\path ([yshift=-\asapTick]A.south); \pgfgetlastxy{\asapbx}{\asapbS}%
\pgfmathsetlengthmacro{\asapband}{\asapbN-\asapbS}%
\pgfmathsetlengthmacro{\asapmazeh}{1.08*\asapband}%
\node[anchor=north west, inner sep=0pt] (M)
    at ([xshift=\asapGm]C3.east|-A.north)
    {\includegraphics[height=\asapmazeh]{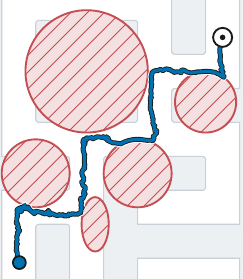}};

\coordinate (groupANW) at ([xshift=-15pt]A.west|-groupNW);
\coordinate (groupASE) at ([xshift=4pt]A.east|-groupSE);
\coordinate (groupACenter) at ($(groupANW)!0.5!(groupASE)$);
\coordinate (groupCNW) at ([xshift=-3pt]M.west|-groupNW);
\coordinate (groupCSE) at ([xshift=3pt]M.east|-groupSE);
\begin{scope}[on background layer]
    \filldraw[groupbox] (groupANW) rectangle (groupASE);
    \filldraw[groupbox] (groupNW) rectangle (groupSE);
    \filldraw[groupbox] (groupCNW) rectangle (groupCSE);
\end{scope}

\node[anchor=south, inner sep=0pt] (LEG)
    at ([yshift=3pt]$(tA.north-|A.west)!0.5!(tA.north-|C3.east)$)
    {\figfont{7}\pgfplotslegendfromname{asapmethods}};
\coordinate (lab) at ([yshift=-11pt]bandS);
\node[plab] at (lab-|groupACenter)  {(a)};
\node[plab] at (lab-|groupCenter) {(b)};
\node[plab] at (lab-|M.south)  {(c)};
\end{tikzpicture}%
\begin{subfigure}{0pt}\phantomsubcaption\label{fig:nfe_score}\end{subfigure}%
\begin{subfigure}{0pt}\phantomsubcaption\label{fig:constraint_levels}\end{subfigure}%
\begin{subfigure}{0pt}\phantomsubcaption\label{fig:multi_constraint}\end{subfigure}%
\par
\captionsetup{justification=justified,singlelinecheck=false,skip=3pt}
\vspace{0.5em}
\caption{\textbf{(a)} \textbf{Task score across sampling steps.} For KUKA block stacking, the score of ASAP and MF passes $1$ at $N=3$ first, but only ASAP is safe. \textbf{(b)} \textbf{Tightening an unseen test-time constraint.}
        For KUKA block stacking, the feasible joint range progressively narrowed. ASAP
        holds safety at every level without retraining, at the lowest cost among safe methods; CoBL,
        Diffuser, CG, and MF violate more as it tightens. Overlapping curves are drawn
        thick-to-thin so each stays visible. \textbf{(c)} \textbf{Five simultaneous constraints.} Two unseen
        Maze2d obstacles become five and ASAP still plans collision-free at
        about $0.06$\,s.}
\label{fig:steps_constraints_conservatism}
\end{figure}

%% file: tables/tables_456.tex
\begin{figure}[t]
\centering
\noindent\mbox{%
\begin{minipage}[t]{0.306\textwidth}\vspace{0pt}
\captionsetup{width=\linewidth,font=asaptables,justification=justified,singlelinecheck=false}
\captionof{table}{\textbf{QP vs.\ post-opt.} Same MeanFlow model and constraints: the affine QP halves runtime with the same score and zero violation.}
    \label{tab:qp_vs_post}
\end{minipage}%
\hspace{0.02\textwidth}%
\begin{minipage}[t]{0.311\textwidth}\vspace{0pt}
\captionsetup{width=\linewidth,font=asaptables,justification=justified,singlelinecheck=false}
\captionof{table}{\textbf{Barrier relaxation.} On Maze2d with $N\!=\!5$, tripling $\bar h$ from $1.0$ to $3.0$ in \cref{eq:discrete_time_varying_relax_term} leaves metrics unaffected.}
    \label{tab:abla_sensitivity}
\end{minipage}%
\hspace{0.02\textwidth}%
\begin{minipage}[t]{0.343\textwidth}\vspace{0pt}
\captionsetup{width=\linewidth,font=asaptables,justification=justified,singlelinecheck=false}
\captionof{table}{\textbf{Adroit Relocate.} In dexterous manipulation, ASAP is the fastest safe method, matching S-Diff. and SAD-F with zero violations.\looseness=-1}
    \label{tab:abla_highdim}
\end{minipage}}
\par\vspace{3pt}
\noindent\mbox{%
\begin{minipage}[t]{0.306\textwidth}\vspace{0pt}\centering
{\fontsize{7.5}{9}\selectfont
    \setlength{\tabcolsep}{2pt}%
    \renewcommand{\arraystretch}{1.10}%
    \begin{tabular*}{\linewidth}{@{\extracolsep{\fill}}lcc@{}}
        \toprule
        \textbf{Maze2d} & \textbf{Post-opt.} & \textbf{ASAP} \\
        \midrule
        s-vio ($\downarrow$)    & 0.000$\pm$0.000 & 0.000$\pm$0.000 \\[3.52pt]
        time ($\downarrow$) & 0.033$\pm$0.015 & \textbf{0.016$\pm$0.006} \\[3.52pt]
        score ($\uparrow$)    & 1.456$\pm$0.467 & 1.456$\pm$0.467 \\[3.52pt]
        \bottomrule
    \end{tabular*}}
\end{minipage}%
\hspace{0.02\textwidth}%
\begin{minipage}[t]{0.311\textwidth}\vspace{0pt}\centering
{\fontsize{7.5}{9}\selectfont
    \setlength{\tabcolsep}{2pt}%
    \renewcommand{\arraystretch}{1.10}%
    \begin{tabular*}{\linewidth}{@{\extracolsep{\fill}}cccc@{}}
        \toprule
        $\bar{h}$ & \textbf{s-vio} ($\downarrow$) & \textbf{time (s)} ($\downarrow$) & \textbf{score} ($\uparrow$) \\
        \midrule
        1.0 & 0.00$\pm$0.00 & 0.25$\pm$0.01 & 1.46$\pm$0.47 \\[3.52pt]
        2.0 & 0.00$\pm$0.00 & 0.25$\pm$0.01 & 1.46$\pm$0.47 \\[3.52pt]
        3.0 & 0.00$\pm$0.00 & 0.25$\pm$0.01 & 1.46$\pm$0.47 \\[3.52pt]
        \bottomrule
    \end{tabular*}}
\end{minipage}%
\hspace{0.02\textwidth}%
\begin{minipage}[t]{0.343\textwidth}\vspace{0pt}\centering
{\fontsize{7.5}{9}\selectfont
    \setlength{\tabcolsep}{2pt}%
    \renewcommand{\arraystretch}{1.10}%
    \begin{tabular*}{\linewidth}{@{\extracolsep{\fill}}lccc@{}}
        \toprule
        \textbf{Method} & \textbf{s-vio} ($\downarrow$) & \textbf{time} ($\downarrow$) & \textbf{score} ($\uparrow$) \\
        \midrule
        MF       & 0.35$\pm$0.07          & \textbf{0.02$\pm$0.00}          & 1.07$\pm$0.12 \\
        S-Diff.  & \textbf{0.00$\pm$0.00} & 0.59$\pm$0.01          & 1.07$\pm$0.08 \\
        SAD-F    & \textbf{0.00$\pm$0.00} & 1.54$\pm$0.15          & 1.05$\pm$0.23 \\
        \textbf{ASAP}
                 & \textbf{0.00$\pm$0.00} & 0.15$\pm$0.00 & \textbf{1.10$\pm$0.06} \\
        \bottomrule
    \end{tabular*}}
\end{minipage}}
\end{figure}

%% file: tables/abla_cbf_projection.tex
\begin{table}[!t]
\centering
\caption{\textbf{Ablation of SDF projection strategies.}
Safety violation (s-vio), planning time (s), and task score
(mean$\pm$std) for final-only SDF projection (Final-only),
full SDF projection at every step (Full), and ASAP on
KUKA stacking ($N=4$) and real-world robot handover ($N=2$). More detail is reported in Appendix~\ref{append:compare_sdf_projection}.}%
\label{tab:abla_cbf_projection}
\vspace{0.3em}
\resizebox{0.8\textwidth}{!}{
\begin{tabular}{
    >{\centering\arraybackslash}m{1.3cm} 
    *{2}{
        >{\centering\arraybackslash}m{1.8cm} 
        *{2}{>{\centering\arraybackslash}m{1.4cm}} 
    }
}
\toprule
& \multicolumn{3}{c}{\textbf{KUKA (Stacking)}} & \multicolumn{3}{c}{\textbf{Real-world robot handover}} \\
\cmidrule(lr){2-4}\cmidrule(lr){5-7}
\textbf{Metric} & \mbox{\textbf{Final-only}} & \textbf{Full} & \textbf{ASAP} & \mbox{\textbf{Final-only}} & \textbf{Full} & \textbf{ASAP} \\
\midrule
s-vio ($\downarrow$) & 0.00$\pm$0.00 & 0.00$\pm$0.00 & 0.00$\pm$0.00 & 0.00$\pm$0.00 & 0.00$\pm$0.00 & 0.00$\pm$0.00 \\
\cmidrule{1-7}
time ($\downarrow$) & 0.05$\pm$0.00 & 0.10$\pm$0.00 & 0.10$\pm$0.00 & 0.02$\pm$0.00 & 0.04$\pm$0.00 & 0.04$\pm$0.01 \\
\cmidrule{1-7}
score ($\uparrow$) & 1.15$\pm$0.79 & 1.45$\pm$0.92 & \textbf{1.75}$\bm{\pm}$\textbf{0.83} & 0.80$\pm$0.40 & 0.70$\pm$0.46 & \textbf{1.00}$\bm{\pm}$\textbf{0.01} \\
\bottomrule\bottomrule
\end{tabular}
}
\vspace{-0.2em}
\end{table}

%% file: tables/tables_abla_robustCBF.tex
\begin{figure}[t]
\centering
\noindent\mbox{%
\begin{minipage}[t]{0.30\textwidth}\vspace{0pt}
\captionsetup{width=\linewidth,font=asaptables,justification=justified,singlelinecheck=false}
\captionof{table}{\textbf{Robust ASAP.} Evaluation of ASAP with robust form \Cref{eq:robust_arb_barrier} across noise levels $\sigma\in\{0.1,0.4,0.8\}$ on Maze2d.}
    \label{tab:robust_CBF}
\end{minipage}%
\hspace{0.02\textwidth}%
\begin{minipage}[t]{0.674\textwidth}\vspace{0pt}
\captionsetup{width=\linewidth,font=asaptables,justification=justified,singlelinecheck=false}
\captionof{table}{\textbf{Comparison of few-step and multi-step safety corrections.} The safety-guaranteed baselines are evaluated with single-step (1) and multi-step (M) safety corrections on Maze2d using the default settings from the original papers or released code. More detail is reported in Appendix~\ref{append:few_step_comparison}.}
    \label{tab:compare_single_multi_step}
\end{minipage}}
\par\vspace{3pt}
\noindent\mbox{%
\begin{minipage}[t]{0.311\textwidth}\vspace{0pt}\centering
{\fontsize{7.1}{9}\selectfont
    \setlength{\tabcolsep}{2pt}%
    \renewcommand{\arraystretch}{1.10}%
    \begin{tabular*}{\linewidth}{@{\extracolsep{\fill}}cccc@{}}
        \toprule
        $\bm{\sigma}$ & \textbf{s-vio} ($\downarrow$) & \textbf{time (s)} ($\downarrow$) & \textbf{score} ($\uparrow$) \\
        \midrule
        0.1 & 0.00$\pm$0.00 & 0.05$\pm$0.01 & 1.42$\pm$0.46 \\[3.52pt]
        0.4 & 0.00$\pm$0.00 & 0.05$\pm$0.01 & 1.41$\pm$0.46 \\[3.52pt]
        0.8 & 0.00$\pm$0.00 & 0.06$\pm$0.01 & 1.41$\pm$0.47 \\[3.52pt]
        \bottomrule
    \end{tabular*}
    }
\end{minipage}%
\hspace{0.02\textwidth}%
\begin{minipage}[t]{0.674\textwidth}\vspace{0pt}\centering
{\fontsize{7.1}{9}\selectfont
    \setlength{\tabcolsep}{1pt}%
    \renewcommand{\arraystretch}{1.10}%
    \begin{tabular*}{\linewidth}{@{\extracolsep{\fill}}lccccccc@{}}
        \toprule
        \textbf{Maze2d} & \textbf{SAD-F (1)} & \textbf{SAD-F (M)} & \textbf{SFM (1)} & \textbf{SFM (M)} & \textbf{S-Diff. (1)} & \textbf{S-Diff. (M)} & \textbf{ASAP (1)}  \\
        \midrule
        s-vio ($\downarrow$) & 0.57$\pm$0.36 & 0.00$\pm$0.00 & 0.05$\pm$0.08 & 0.00$\pm$0.00 & 0.71$\pm$0.20 & 0.00$\pm$0.00 & 0.00$\pm$0.00 \\[3.52pt]
        time ($\downarrow$) & 0.10$\pm$0.01 & 35.84$\pm$2.51 & 0.11$\pm$0.00 & 15.50$\pm$0.75 & 0.86$\pm$0.18 & 17.08$\pm$2.29 & 0.06$\pm$0.01 \\[3.52pt]
        score ($\uparrow$) & 1.61$\pm$0.07 & 1.44$\pm$0.28 & 0.76$\pm$0.66 & 1.30$\pm$0.67 & 0.80$\pm$0.57 & 1.46$\pm$0.23 & 1.46$\pm$0.47 \\[3.52pt]
        \bottomrule
    \end{tabular*}
    }
\end{minipage}}
\vspace{-0.5em}
\end{figure}

%% file: sections/conclusion.tex
\section{Conclusion}
We proposed ASAP, a generative planner with formal safety guarantees for arbitrary-step generation. By integrating an affine barrier constraint into MeanFlow and adapting the safety correction to each sampling interval, ASAP enables fast trajectory generation while enforcing test-time safety constraints without retraining. Across navigation, locomotion, and manipulation tasks, ASAP satisfies every safety constraint while planning at least $20\times$ faster than safety-guaranteed baselines and maintaining competitive task performance; For real-world robot handover task, it remains safe and $200\times$ faster. Future work will extend the framework beyond current assumptions to dynamic consistency and richer constraint geometries, broadening its applicability to complex robotic systems.

%% file: sections/appendix.tex
\clearpage
\appendix
\onecolumn

\mtcaddpart[Appendix]
\par\vspace*{1.5ex}
{\raggedright\normalfont\Large Appendix\par}
\nobreak
\vskip 0.75em
\parttoc

\section{Signed Distance Functions} \label{append:SDF}

Given a constraint set $\mathbb{S}$ with boundary $\partial\mathbb{S}$, a signed distance function (SDF) measures the shortest distance to the boundary, while its sign indicates whether the constraint is satisfied:
\begin{align}\label{eq:sdf}
    h(\bm{\tau}_{t_i})=\begin{cases}
        \min_{\bm{p}\in\partial\mathbb{S}} ||\bm{\tau}_{t_i}\!-\!\bm{p}||&\text{if } \bm{\tau}_{t_i}\in\mathbb{S}\\
        -\min_{\bm{p}\in\partial\mathbb{S}} ||\bm{\tau}_{t_i}\!-\!\bm{p}||&\text{else }
    \end{cases}
\end{align}
Accordingly, $h(\bm{x})\geq0$ indicates constraint satisfaction. In trajectory planning, this SDF can be evaluated at each trajectory state $\bm{\tau}_{t_i}(k)$ to obtain the state-wise barrier $h(\bm{\tau}_{t_i}(k))$.

\paragraph{Multiple constraints as a single SDF.}
Multiple obstacle constraints can be represented by a single SDF by first combining the unsafe sets. Let $\{\mathbb{O}_j\}_{j=1}^M$ denote $M$ unsafe obstacle sets. The total unsafe region is
\begin{equation}\label{eq:multiple_obs}
    \mathbb{O}=\bigcup_{j=1}^M \mathbb{O}_j,
\end{equation}
and the corresponding safe set is
\begin{equation} \label{eq:multiple_obs_safe_Set}
    \mathbb{S}=\mathbb{R}^d\setminus \mathbb{O}.
\end{equation}
The SDF in \cref{eq:sdf} can then be defined directly with respect to this combined safe set $\mathbb{S}$. Thus, even with multiple obstacles, safety can still be expressed through a single scalar condition
\begin{equation}
    h(\bm{x})\ge0. \label{eq:multiple_obs_sdf}
\end{equation}

For common obstacle-avoidance constraints, this combined SDF has a simple form. If each obstacle $\mathbb{O}_j$ has an individual SDF $h_j(\bm{x})$ with positive value outside the obstacle, then for strictly separated obstacles the combined SDF is
\begin{equation}
    h(\bm{x})=\min_{j=1,\dots,M} h_j(\bm{x}).
\end{equation}
Equivalently, this follows from the distance to a union if points are outside the obstacle union:
\begin{equation}
    \operatorname{dist}\!\left(\bm{x},\bigcup_{j=1}^M \mathbb{O}_j\right)
    =
    \min_{j=1,\dots,M}\operatorname{dist}(\bm{x},\mathbb{O}_j).
\end{equation}
For overlapping obstacles, the minimum composition still characterizes safety but may not equal the exact union SDF inside overlapping regions. In such cases, the exact SDF is defined directly from the combined unsafe region in \cref{eq:multiple_obs}--\cref{eq:multiple_obs_sdf}.

Therefore, the combined SDF automatically selects the closest active obstacle. At a differentiable point where the nearest obstacle is unique, say
\begin{equation}
    j^\star=\arg\min_{j} h_j(\bm{x}),
\end{equation}
we have
\begin{equation}
    h(\bm{x})=h_{j^\star}(\bm{x}),
    \qquad
    \nabla h(\bm{x})=\nabla h_{j^\star}(\bm{x}).
\end{equation}
Thus, the SDF gradient points toward the nearest active safety boundary, which is the direction used by the minimum-norm affine barrier correction in our method.

At points where multiple obstacles are equally close, the minimum composition may be non-differentiable. Such points correspond to ties between nearest boundaries. In practice, they can be handled by generalized gradients, smoothing, or active-boundary selection, as discussed in \Cref{sec:arb_safe_flow}. These cases do not change the sign condition $h(\bm{x})\ge0$, and the differentiable case covers almost all regular points.

\paragraph{Examples.}
For circular obstacles $\mathbb{O}_j=\{\bm{x}:\|\bm{x}-\bm{c}_j\|\le r_j\}$, the individual SDF is
\begin{equation}
    h_j(\bm{x})=\|\bm{x}-\bm{c}_j\|-r_j.
\end{equation}
The combined SDF for avoiding all obstacles is
\begin{equation}
    h(\bm{x})=\min_{j=1,\dots,M}\left(\|\bm{x}-\bm{c}_j\|-r_j\right).
\end{equation}
Then $h(\bm{x})\ge0$ holds if and only if $\bm{x}$ is outside every obstacle.

Equivalent barrier functions that preserve the same zero level set and sign can also be used. For instance, collision avoidance outside a circle can be represented by $h(x,y)=(x-x_0)^2+(y-y_0)^2-r^2\ge0$ \citep{origane2025risk}, and more generally, an ellipsoidal constraint can be expressed through a quadratic barrier of the form $h(\bm{x})=(\bm{x}-\bm{c})^\top Q(\bm{x}-\bm{c})-1\ge0$.


\section{Proofs of Theoretical Results} \label{append:proof}

\subsection{Proof of \cref{lemma:PTCBF-Flow}}
\begin{proof}[Proof of \Cref{lemma:PTCBF-Flow}] The time-varying CBF in \cref{eq:QP_continuous_CBF} is equilavant to $\dot{B}+B\ge0$, where $B:=B_t$ and $B_t=h(\bm{\tau}_t)-\rho_t$. The initial barrier function $B_{0}\ge0$ due to our design of $\rho_0$, based on the scheduling design of $\rho_t$, by solving the QP in \cref{eq:QP_continuous_CBF}, , it will remain in the positive region for the full time $B_t\ge0,t\in[0,1]$ with control barrier function theory \citep{ames2016control}. Since $\rho_1=0$, the barrier function $B_1=h(\bm{\tau}_1)\geq0$, which implies the safe trajectory generation. This formulation is similar to the prescribed-time control barrier function \citep{abel2023prescribed, huang2024learning}, which designs a time-varying CBF to guarantee safety constraints at a predefined time. 
\end{proof}

\subsection{Proof of \cref{th:arb_Step_safe_flow}}
\begin{proof}[Proof of \Cref{th:arb_Step_safe_flow}]
    If $h(\hat{\bm{\tau}}_{t_i})\geq\rho_{t_{i+1}}$, the minimum-norm QP returns $\bm{u}_{t_i}=\bm{0}$, so the next-step condition holds. Otherwise, $h(\hat{\bm{\tau}}_{t_i})<\rho_{t_{i+1}}\leq0$. The exact SDF has a unit gradient at this unsafe nominal point, and the minimum-norm solution is~\citep{ames2016control}
    \begin{equation} \label{eq:minimum_correct_SDF}
        \bm{u}_{t_i}\Delta t
        =\frac{\bigl(\rho_{t_{i+1}}-h(\hat{\bm{\tau}}_{t_i})\bigr)}{\|\nabla h(\hat{\bm{\tau}}_{t_i})\|^2}\nabla h(\hat{\bm{\tau}}_{t_i})
        =\bigl(\rho_{t_{i+1}}-h(\hat{\bm{\tau}}_{t_i})\bigr)\nabla h(\hat{\bm{\tau}}_{t_i}).
    \end{equation}

    Let $\bm{p}\in\partial\mathbb{S}$ be a nearest safe point to $\hat{\bm{\tau}}_{t_i}$. Differentiability of the exact SDF implies $\bm{p}-\hat{\bm{\tau}}_{t_i}=-h(\hat{\bm{\tau}}_{t_i})\nabla h(\hat{\bm{\tau}}_{t_i})$. For a displacement of length $0\leq s\leq-h(\hat{\bm{\tau}}_{t_i})$ along this gradient, the triangle inequality gives the lower bound below, while the candidate $\bm{p}\in\mathbb{S}$ gives the upper bound:
    \begin{align*}
        \operatorname{dist}\bigl(\hat{\bm{\tau}}_{t_i}+s\nabla h(\hat{\bm{\tau}}_{t_i}),\mathbb{S}\bigr)
        &\geq\operatorname{dist}(\hat{\bm{\tau}}_{t_i},\mathbb{S})-s
        =-h(\hat{\bm{\tau}}_{t_i})-s,\\
        \operatorname{dist}\bigl(\hat{\bm{\tau}}_{t_i}+s\nabla h(\hat{\bm{\tau}}_{t_i}),\mathbb{S}\bigr)
        &\leq\bigl\|\hat{\bm{\tau}}_{t_i}+s\nabla h(\hat{\bm{\tau}}_{t_i})-\bm{p}\bigr\|
        =-h(\hat{\bm{\tau}}_{t_i})-s.
    \end{align*}
    The two bounds coincide. The corrected point remains outside $\mathbb{S}$ until it reaches $\bm{p}$, so its signed distance satisfies
    \begin{equation} \label{eq:sdf_normal_segment}
        h\bigl(\hat{\bm{\tau}}_{t_i}+s\nabla h(\hat{\bm{\tau}}_{t_i})\bigr)
        =h(\hat{\bm{\tau}}_{t_i})+s,
        \qquad 0\leq s\leq-h(\hat{\bm{\tau}}_{t_i}).
    \end{equation}

    Since $0<\rho_{t_{i+1}}-h(\hat{\bm{\tau}}_{t_i})\leq-h(\hat{\bm{\tau}}_{t_i})$, the optimal correction stays on this segment that $s^*=\rho_{t_{i+1}}-h(\hat{\bm{\tau}}_{t_i})$. With $\bm{\tau}_{t_{i+1}}=\hat{\bm{\tau}}_{i}+s^*\nabla h(\hat{\bm{\tau}}_{t_i})$ induced by \cref{eq:minimum_correct_SDF} and our update rule in \cref{eq:update_control}, we can obtain the following equation
    \begin{equation} \label{eq:proof_arb_step_barrier_function}
        h(\bm{\tau}_{t_{i+1}})
        =
        h\bigl(\hat{\bm{\tau}}_{t_i}+s^*\nabla h(\hat{\bm{\tau}}_{t_i})\bigr)
        =
        h(\hat{\bm{\tau}}_{t_i})+s^*
        =
        \rho_{t_{i+1}},
        \qquad i=0,\ldots,N-1.
    \end{equation}
    At the final step, $\rho_{t_N}=0$: an unsafe nominal state is corrected exactly to the nearest safe boundary, while an already safe nominal state is unchanged. Applying this argument to every trajectory state proves terminal safety for any $N\geq1$. The terminal sample is $\bm{\tau}_{t_N}$ in discrete-step notation, corresponding to $\bm{\tau}_1$ at generative time $t=1$.
\end{proof}
The assumption of \Cref{th:arb_Step_safe_flow} requires the feasibility during QP solving. Even though it is a well-known topic of the standard CBF-based method, it is typically not an issue for our case due to the fully actuated, integrator-like generation dynamics in \cref{eq:update_control}. We will discuss the details of feasibility in \Cref{th:CBF_feasibility}.

\subsection{Proof of \cref{corollary:one_step_safe_flow}}
\begin{proof}[Proof of \Cref{corollary:one_step_safe_flow}] Holding the affine barrier function \cref{eq:arb_Step_barrier_function}, we can directly obtain the following inequality by setting $N=1$ for \Cref{th:arb_Step_safe_flow}:
\begin{equation}
    h(\bm{\tau}_{1}) = 
        h(\hat{\bm{\tau}}_0) +
        \nabla h(\hat{\bm{\tau}}_0)^\top(\bm{u}_0\Delta t)
        \ge
        \rho_{1}, 
        \quad
        \Delta t=1
\end{equation}
Since $\rho_1=0$, we can prove that the trajectory $\bm{\tau}_1$ is safe at $t=1$ in one step generation.
\end{proof}
The \Cref{corollary:one_step_safe_flow} can be viewed as a special case of \Cref{th:arb_Step_safe_flow}. When $N=1$, it shows that terminal safety can still be guaranteed with only a single generative update and one safety correction, thereby enabling one-step safe trajectory generation.

\subsection{Proof of \cref{prop:robust_safe_flow}}
\begin{proof}[Proof of \Cref{prop:robust_safe_flow}]
    Consider an arbitrary sampling step $i$. Let
    \begin{equation}
        \hat H_i =
        \hat h(\hat{\bm{\tau}}_{t_i})
        +\hat{\bm{g}}(\hat{\bm{\tau}}_{t_i})^\top
        (\bm{u}_{t_i}\Delta t)
    \end{equation}
    denote the computed affine barrier, and let
    $\epsilon_i^{\mathrm{QP}}=
    \max\{0,\rho_{t_{i+1}}+r_{t_{i+1}}-\hat H_i\}$
    denote the numerical constraint violation. Assume the integrated barrier error is bounded by $\bar e$:
    \begin{equation}\label{eq:integrated_barrier_error}
        |\hat H_i-h(\bm{\tau}_{t_{i+1}})|
        +\epsilon_i^{\mathrm{QP}}
        \leq \bar e.
    \end{equation}
    By the robust affine constraint in
    \cref{eq:robust_arb_barrier}, we have
    $\hat H_i\geq\rho_{t_{i+1}}+r_{t_{i+1}}
    -\epsilon_i^{\mathrm{QP}}$.
    Therefore,
    \begin{align}
        h(\bm{\tau}_{t_{i+1}})
        &\geq \hat H_i
        -|\hat H_i-h(\bm{\tau}_{t_{i+1}})| \nonumber\\
        &\geq \rho_{t_{i+1}}+r_{t_{i+1}}
        -\epsilon_i^{\mathrm{QP}}
        -|\hat H_i-h(\bm{\tau}_{t_{i+1}})| \nonumber\\
        &\geq \rho_{t_{i+1}}+r_{t_{i+1}}-\bar e.
    \end{align}
    At the terminal step, $r_{t_N}=\bar e$
    and $\rho_{t_N}=0$, yielding
    \begin{equation}
        h(\bm{\tau}_{t_N})\geq
        \rho_{t_N}+r_{t_N}-\bar e=0.
    \end{equation}
    Thus, the terminal trajectory is safe for any
    $N\geq1$, provided the robust QP is feasible and
    the integrated error bound holds.
\end{proof}

\subsection{Proof of \cref{th:zero_conservatism}} \label{append:proof_conservatism}
\begin{proof}[Proof of \Cref{th:zero_conservatism}]
Consider an arbitrary sampling step $i$ and denote
$$
    \bm{g}_{t_i}:=\nabla h(\hat{\bm{\tau}}_{t_i}),
    \qquad
    b_{t_{i}}:=\rho_{t_{i+1}}-h(\hat{\bm{\tau}}_{t_i}).
$$
The affine barrier constraint in \cref{eq:arb_Step_barrier_function} becomes
$$
    \bm{g}_{t_i}^\top(\bm{u}_{t_i}\Delta t)\geq b_{t_{i}}.
$$
If $b_{t_{i}}\leq0$, the nominal sample already satisfies the affine barrier constraint. Hence, the minimum-norm QP in \cref{eq:QP_arb_Step_barrier_function} returns $\bm{u}_{t_i}=\bm{0}$, such that $\bm{\tau}_{t_{i+1}}=\hat{\bm{\tau}}_{t_i}$. Therefore, the barrier value equals the true next-step barrier value and $C_i=0$. If $b_{t_{i}}>0$, the barrier constraint is active. The minimum-norm solution aligns with $\bm{g}_{t_i}$, since any component orthogonal to $\bm{g}_{t_i}$ increases the objective without affecting the constraint. Enforcing the active constraint with equality gives
$$
    \bm{u}_{t_i}\Delta t
    =
    \frac{b_{t_{i}}}{\|\bm{g}_{t_i}\|^2}\bm{g}_{t_i}.
$$
Since $h$ is an exact differentiable SDF and $b_{t_{i}}>0$ implies $h(\hat{\bm{\tau}}_{t_i})<\rho_{t_{i+1}}\leq0$, $\hat{\bm{\tau}}_{t_i}$ lies outside the safe set and $||\bm{g}_{t_i}||=1$. Moreover, $\bm{g}_{t_i}$ points along the shortest direction toward the nearest safe boundary. Let $s\!\geq\!0$ denote the traveled distance along this unit direction. The SDF therefore satisfies\looseness=-1
\begin{equation} \label{eq:exact_SDF_property}
    h(\hat{\bm{\tau}}_{t_i}+s\bm{g}_{t_i})
    =
    h(\hat{\bm{\tau}}_{t_i})+s,
    \qquad
    0\leq s\leq-h(\hat{\bm{\tau}}_{t_i}).
\end{equation}
Because $b_{t_{i}}=\rho_{t_{i+1}}-h(\hat{\bm{\tau}}_{t_i})\leq-h(\hat{\bm{\tau}}_{t_i})$, the QP correction remains within this interval. Using $||\bm{g}_{t_i}||=1$, we have $\bm{u}_{t_i}\Delta t=b_{t_{i}}\bm{g}_{t_i}$. Following the update rule \cref{eq:update_control} that $\bm{\tau}_{t_{i+1}}=\hat{\bm{\tau}}_{t_i}+\bm{u}_{t_i}\Delta t$, the exact value of SDF at next step $h(\bm{\tau}_{t_{i+1}})$ is $h(\bm{\tau}_{t_{i+1}})
    =h(\hat{\bm{\tau}}_{t_i}+\bm{u}_{t_i}\Delta t)
    =
    h(\hat{\bm{\tau}}_{t_i}+b_{t_{i}}\bm{g}_{t_i})$. Based on the \cref{eq:exact_SDF_property}, we can derive this exact value as follows
\begin{equation}
    h(\bm{\tau}_{t_{i+1}})=
    h(\hat{\bm{\tau}}_{t_i}+b_{t_{i}}\bm{g}_{t_i})=
    h(\hat{\bm{\tau}}_{t_i})+b_{t_{i}}
    =
    \rho_{t_{i+1}}.
\end{equation}
The affine formulation from \cref{eq:arb_Step_barrier_function} yields the same value:
\begin{equation}
    \hat{h}(\hat{\bm{\tau}}_{t_i})
    =
    h(\hat{\bm{\tau}}_{t_i})
    +
    \bm{g}_{t_i}^\top(\bm{u}_{t_i}\Delta t)=h(\hat{\bm{\tau}}_{t_i})
    + \bm{g}_{t_i}^\top\bm{g}_ib_{t_{i}}=
    h(\hat{\bm{\tau}}_{t_i})+b_{t_{i}}
    =
    \rho_{t_{i+1}}.
\end{equation}
Thus, its barrier value and the true next-step barrier value coincide, i.e.,
$
h(\bm{\tau}_{t_{i+1}})=\hat{h}(\hat{\bm{\tau}}_{t_i}),
$
which gives $C_i=0$ by \cref{eq:conservatism_term}. Since this holds for every $i=0,\ldots,N-1$, $\bar{C}=0$ for arbitrary $N$.
\end{proof}

\subsection{Feasibility of Arbitrary-Step Affine Barrier Constraints}
\label{append:proof_feasibility}

\begin{theorem}\label{th:CBF_feasibility}
Suppose $h(\cdot)$ is differentiable for each $\hat{\bm{\tau}}_{t_i}$,,
then, for every generation step $i$, there exists a correction
$\bm{u}_{t_i}$ satisfying the affine barrier constraint
\cref{eq:arb_Step_barrier_function}.
\end{theorem}

\begin{proof}
The affine barrier constraint can be written as
\begin{equation}
    \nabla h(\hat{\bm{\tau}}_{t_i})^\top
    (\bm{u}_{t_i}\Delta t)
    \geq
    \rho_{t_{i+1}}-h(\hat{\bm{\tau}}_{t_i}).
\end{equation}
If $h(\hat{\bm{\tau}}_{t_i})\geq\rho_{t_{i+1}}$, the constraint is already
satisfied by $\bm{u}_{t_i}=\bm{0}$. Otherwise, a feasible
correction can always be chosen along the barrier gradient, e.g.,
\begin{equation}
    \bm{u}_{t_i}
    =
    \frac{\rho_{t_{i+1}}-h(\hat{\bm{\tau}}_{t_i})}
    {\Delta t\,\|\nabla h(\hat{\bm{\tau}}_{t_i})\|^2}
    \nabla h(\hat{\bm{\tau}}_{t_i}).
\end{equation}
This choice satisfies the affine barrier constraint with equality,
establishing feasibility at every step.
\end{proof}

For an exact signed distance function (SDF), differentiability implies
$\|\nabla h\|=1$ wherever the SDF is differentiable, so the
non-vanishing-gradient condition in \cref{th:CBF_feasibility} is
naturally satisfied. SDFs may nevertheless be non-differentiable at
certain points, although they are differentiable almost everywhere
under standard regularity conditions on the underlying set
\citep{gilbarg1977elliptic}. In practice, such non-smoothness can be
handled through smooth approximations \citep{begzadic2025learning} or
by decomposing a complex constraint set into multiple subsets and
constructing a separate barrier for each subset.

For example, the commonly used box constraint
$\|\bm{\tau}_{t_i}(k)\|_\infty\leq\bar{\tau}$ can be decomposed into
individual upper and lower bounds, yielding affine barriers of the form
$h_k(\bm{\tau}_{t_i})=\bar{\tau}\pm\bm{\tau}_{t_i}(k)$.
These barriers are differentiable with constant nonzero gradients.
Consequently, the regularity condition required by
\cref{th:CBF_feasibility} is readily satisfied for many practical
constraint constructions used in CBF-based planning
\citep{long2021learning}. \looseness=-1

\subsection{Extension to General Barrier Functions}
\label{append:general_barrier}
In the main paper, we instantiate the barrier function $h$ as an exact signed distance function (SDF), which provides a natural representation of geometric safety constraints and informative gradients toward the safe set \citep{long2021learning,park2019deepsdf}. Importantly, its geometric properties enable the safety guarantee in \Cref{th:arb_Step_safe_flow} without requiring convexity of $h$. Nevertheless, ASAP is not restricted to exact SDFs. We next show that arbitrary-step safety can also be guaranteed for a general differentiable barrier function under a convexity assumption.
\begin{theorem}[\textbf{Arbitrary-Step Safety with Convex Barriers}]
\label{th}
If $h$ is differentiable and convex, the control input $\bm{u}_{t_i}$ given by \cref{eq:QP_arb_Step_barrier_function} ensures that the solution $\bm{\tau}_{t_{N}}$ of \cref{eq:update_control} satisfies the safety constraint $h(\bm{\tau}_{t_{N}})\geq0$ for any initial condition $\bm{\tau}_0$, 
\end{theorem}
\begin{proof}
By convexity of $h$, its first-order approximation provides a global lower bound:
\begin{align}
h(\bm{\tau}_{t_{i+1}})=h(\hat{\bm{\tau}}_{t_i}+\bm{u}_{t_i}\Delta t)\geq
h(\hat{\bm{\tau}}_{t_i})
+\nabla h(\hat{\bm{\tau}}_{t_i})^\top(\bm{u}_{t_i}\Delta t)
\geq\rho_{t_{i+1}},
\label{eq}
\end{align}
where the second inequality follows from \cref{eq:arb_Step_barrier_function}. Hence, $h(\bm{\tau}_{t_{i+1}})\geq\rho_{t_{i+1}}$ for every $i=0,\ldots,N-1$. Since $\rho_{t_N}=0$, the terminal sample $\bm{\tau}_{t_N}$, corresponding to $\bm{\tau}_1$ at generative time $t=1$, satisfies $h(\bm{\tau}_1)\geq0$.
\end{proof}
\begin{remark}
    The convexity assumption is imposed on the certificate function $h(\cdot)$, rather than on its safe superlevel set $\mathcal{C}=\{\bm{\tau}: h(\bm{\tau}) \geq 0\}$, and therefore does not restrict the safe region to be convex. In particular, commonly used distance-based collision-avoidance barriers, such as $h(p)=||p-c||^2-r^2$, for spherical obstacles \citep{grover2023before}, are convex, while their corresponding collision-free sets are nonconvex. The same property extends naturally to ellipsoidal obstacles through quadratic barriers $h(p)=(p-c)^TQ(p-c)-1, Q\succ 0$, which are closely related to ellipsoidal CBF constructions widely used in robotic collision avoidance \citep{funada2024collision, hu2025vlsa, huang2026intervention}. Hence, the convexity assumption accommodates a broad class of practically relevant geometric safety constraints. 
\end{remark}
\paragraph{Relaxation to Nonconvex Barrier Functions.}
The convexity requirement can be further relaxed when the negative curvature of $h$ is bounded. Specifically, assume that $h$ is twice differentiable and
\begin{equation}
\nabla^2 h(\bm{\tau})\succeq-\mu\bm{I},
\qquad \mu\in\mathbb{R}_+,
\end{equation}
over the relevant domain, and an upper bound of control $|\bm{u}_{t_i}|\leq\bar U$, where we discuss at the end of this section. We obtain the following result.
\begin{theorem}[\textbf{Safety with Bounded Nonconvexity}]
\label{th:h_relax}
Under the assumptions above, the solution $\bm{\tau}_{t_N}$ of \cref{eq:update_control} is safe if the control input $\bm{u}_{t_i}$ satisfies
\begin{equation}
\label{eq:barrier_function_nonconvex}
h(\hat{\bm{\tau}}_{t_i})
+\nabla h(\hat{\bm{\tau}}_{t_i})^\top(\bm{u}_{t_i}\Delta t)
\geq
\rho_{t_{i+1}}
+\frac{\mu\bar U^2}{2N^2}.
\end{equation}
\end{theorem}
\begin{proof}
Let $\bm{d}_{t_i}=\bm{u}_{t_i}\Delta t$. By Taylor's theorem, for some $\xi\in[0,1]$,
\begin{align} \label{eq:second_order_taylor}
h(\bm{\tau}_{t_{i+1}})
=h(\hat{\bm{\tau}}_{t_i})
+\nabla h(\hat{\bm{\tau}}_{t_i})^\top\bm{d}_{t_i} +\frac{1}{2}\bm{d}_{t_i}^\top
\nabla^2h(\hat{\bm{\tau}}_{t_i}+\xi\bm{d}_{t_i})\bm{d}_{t_i} .
\end{align}
The bounded-curvature assumption implies
\begin{equation} \label{eq:taylor_low_curvature_ineq}
\frac{1}{2}\bm{d}_{t_i}^\top
\nabla^2h(\hat{\bm{\tau}}_{t_i}+\xi\bm{d}_{t_i})\bm{d}_{t_i}
\geq-\frac{\mu}{2}||\bm{d}_{t_i}||^2.
\end{equation}
Since $|\bm{u}_{t_i}|\leq\bar U$ (see the discussion at the end of this section) and $\Delta t=1/N$, the right term in \cref{eq:taylor_low_curvature_ineq} follows $-\frac{\mu}{2}||\bm{d}_{t_i}||^2 \ge -\frac{\mu\bar{U}^2}{2N^2}$ and we can rewrite \cref{eq:second_order_taylor} as follow
\begin{equation}
h(\bm{\tau}_{t_{i+1}})
\geq
h(\hat{\bm{\tau}}_{t_i})
+\nabla h(\hat{\bm{\tau}}_{t_i})^\top(\bm{u}_{t_i}\Delta t)
-\frac{\mu\bar U^2}{2N^2}.
\end{equation}
If the robust barrier function in \cref{eq:barrier_function_nonconvex} is satisfied, inducing $h(\hat{\bm{\tau}}_{t_i})\!+\!\nabla h(\hat{\bm{\tau}}_{t_i})^\top\!(\bm{u}_{t_i}\Delta t)\!-\!\frac{\mu\bar{U}^2}{2N^2} \!\ge\! \rho_{t_{i+1}}$, we can prove that $\bm{\tau}_{t_N}$ is safe since $h(\bm{\tau}_{t_{i+1}}) \!\ge\! h(\hat{\bm{\tau}}_{t_i})+
        \nabla h(\hat{\bm{\tau}}_{t_i})^\top(\bm{u}_{t_i}\Delta t)
        -
        \frac{\mu\bar{U}^2}{2N^2} \ge \rho_{t_{i+1}}$.
Therefore, satisfying \cref{eq:barrier_function_nonconvex} yields
$h(\bm{\tau}_{t_{i+1}})\!\geq\!\rho_{t_{i+1}}$ for every step $i=0,\cdots,N-1$. Since $\rho_{t_{N}}\!=\!0$, terminal safety follows.
\end{proof}
The additional term $\frac{\mu\bar U^2}{2N^2}$ acts as a robust margin accounting for the bounded negative curvature of $h$. With \Cref{th:h_relax}, we can apply our ASAP framework with a robust barrier function as defined in \cref{eq:barrier_function_nonconvex} and keep an efficient QP formulation even if $h(\cdot)$ is non-convex. For $\mu=0$, $h$ is convex and \cref{eq:barrier_function_nonconvex} recovers the original affine barrier constraint in \cref{eq:arb_Step_barrier_function}.

For the conservatism analysis of the general CBF, we derive it as follows.

\begin{theorem}[\textbf{Arbitrary-Step Conservatism Bound for general CBF}] \label{th:conservative_bound_general_CBF}
If $h(\cdot)$ is $L$-smooth, i.e., $||\nabla^2 h(\bm{\tau})||_2\leq L$ and convex $||\nabla^2 h(\cdot)||\ge0$, where $L$ is the Lipschitz constant of $\nabla h$, the conservatism $\bar{C}$ induced by \cref{eq:QP_arb_Step_barrier_function} is bounded and decreases with the number of sampling steps $N$ as\looseness=-1
\begin{equation} \label{eq:conservative_bound_general_CBF}
    \bar{C} \le\frac{L\bar{U}^2}{2N}, \quad \text{where } \bar{C}=O(1/N)
\end{equation}
\end{theorem}
This theorem shows that the conservatism induced by the finite-step affine approximation remains bounded even for small $N$. As $N\rightarrow\infty$ and $\Delta t\rightarrow0$, the approximation-induced conservatism vanishes, recovering the continuous-time barrier evaluation. Thus, our analysis characterizes the trade-off between sampling efficiency and conservatism through the choice of $N$.

\begin{remark}
The $L$-smoothness assumption is mild for SDF-based constraints. Although exact SDFs can be non-smooth at certain points, they are smooth with Lipschitz-continuous gradients over regular regions of the planning domain. Moreover, SDFs can be smoothly approximated \citep{begzadic2025learning} or constructed over simpler decomposed constraint sets to obtain $L$-smooth barriers. Thus, $L$-smoothness can be readily satisfied for practical SDF-based safety constraints.
\end{remark}

\begin{proof}[Proof of \Cref{th:conservative_bound_general_CBF}] The constraint residual is defined for each step as
\begin{equation}
    C_i
    =
    h(\bm{\tau}_{t_{i+1}})
    -
    h(\hat{\bm{\tau}}_{t_i})
    -
    \nabla h(\hat{\bm{\tau}}_{t_i})^\top(\bm{u}_{t_i}\Delta t), 
    \quad
    \forall i = 0,\ldots,N-1,
\end{equation}
where the first-order approximation is $\hat{h}(\hat{\bm{\tau}}_{t_i})=h(\hat{\bm{\tau}}_{t_i})+\nabla h(\hat{\bm{\tau}}_{t_i})^\top(\bm{u}_{t_i}\Delta t)$. By Taylor's theorem, we can obtain the exact barrier value of the next step $h(\bm{\tau}_{t_{i+1}})$ as
\begin{equation} \label{eq:taylor}
    h(\bm{\tau}_{t_{i+1}})=h(\hat{\bm{\tau}}_{t_i})+\nabla h(\hat{\bm{\tau}}_{t_i})^\top(\bm{u}_{t_i}\Delta t)+\frac{1}{2}(\bm{u}_{t_i}\Delta t)^\top \nabla^2h(\hat{\bm{\tau}}_{t_i}+\xi\bm{u}_{t_i}\Delta t)(\bm{u}_{t_i}\Delta t),
\end{equation}
where the second-order prediction has an upper bound if $h(\cdot)$ is $L$-smooth
\begin{equation} \label{eq:taylor_upper_bound}
    (\bm{u}_{t_i}\Delta t)^\top \nabla^2h(\hat{\bm{\tau}}_{t_i}+\xi\bm{u}_{t_i}\Delta t)(\bm{u}_{t_i}\Delta t) 
    \le
    (\bm{u}_{t_i}\Delta t)^\top LI(\bm{u}_{t_i}\Delta t)
    =
    L||\bm{u}_{t_i}\Delta t||^2.
\end{equation}
By substituting \cref{eq:taylor_upper_bound} to \cref{eq:taylor}, we can obtain the inequality as follows:
\begin{equation}
    h(\bm{\tau}_{t_{i+1}})
    -
    h(\hat{\bm{\tau}}_{t_i})
    -
    \nabla h(\hat{\bm{\tau}}_{t_i})^\top(\bm{u}_{t_i}\Delta t)
    \le
    \frac{L}{2}||\bm{u}_{t_i}\Delta t||^2.
\end{equation}
As a result, we can derive the upper bound of the constraint residual at each step
\begin{equation}
    C_i
    \le
    \frac{L}{2}||\bm{u}_{t_i}\Delta t||^2
\end{equation}
and the total conservatism is shown as follows
\begin{equation}
    \bar{C}
    =
    \sum_{i=0}^{N-1}C_i
    \le
    \frac{L}{2}
    \sum_{i=0}^{N-1}
    \|\bm{u}_{t_i}\Delta t\|^2.
\end{equation}
Since the $||\bm{u}_{t_i}||\le\bar{U}$ and $\Delta t=\frac{1}{N}$, we can prove that $\bar{C} \le \frac{L\bar{U}^2}{2N}$.
\end{proof}

\textbf{How to estimate the value of $\bar{U}$?} For our ASAP framework, we calculate $\bm{u}_{t_i}$ by \cref{eq:QP_arb_Step_barrier_function} to minimally perturb the MeanFlow update with following constraint
\begin{equation} \label{eq:barrier_const_deficit}
    \nabla h(\hat{\bm{\tau}}_{t_i})^\top \bm{u}_{t_i}\Delta t \ge d_i,
\end{equation}
where the $d_i$ is the scheduled deficit $d_i=\rho_{t_{i+1}}-h(\hat{\bm{\tau}}_{t_i})=\rho_{t_{i+1}}-\rho_{t_i}-(h(\hat{\bm{\tau}}_{t_i})-\rho_{t_i})$. Based on the time-varying scheduled term $\rho_{t_i}$ defined in \cref{eq:discrete_time_varying_relax_term}, we can derive a weighting term $w_i$ that schedule the constraint level with the range $[-\bar{h},0]$ evenly as follow
\begin{equation} \label{eq:weighting_schedule}
    w_i=\frac{(\rho_{t_{i+1}}-\rho_{t_i})}{\bar{h}}=\frac{1}{N}.
\end{equation}
For the deviation between $h(\hat{\bm{\tau}}_{t_i})$ and $\rho_{t_i}$, we can obtain the following inequality if $h(\cdot)$ is Lipschitz, which is a mild assumption in real-world applications \citep{grover2023before}
\begin{equation} \label{eq:lipschitz_h}
    \|h(\hat{\bm{\tau}}_{t_i})-h(\bm{\tau}_{t_i})\|
    \le
    G||\bm{\mu}_{t_i}^{\bm{\theta}}||\Delta t,
\end{equation}
where the $G\in \mathbb{R}_+$ is the Lipschitz constant of the $h(\cdot)$ and $\Delta t=\frac{1}{N}$. By reformulating \cref{eq:lipschitz_h} as $h(\hat{\bm{\tau}}_{t_i})-h(\bm{\tau}_{t_i}) \ge -G||\bm{\mu}_{t_i}^{\bm{\theta}}||\Delta t$ by Absolute Value Inequalities, we can derive 
\begin{equation} \label{eq:ineq_h_rho}
    h(\hat{\bm{\tau}}_{t_i}) \ge h(\bm{\tau}_{t_i})-\frac{G||\bm{\mu}_{t_i}^{\bm{\theta}}||}{N}\ge\rho_{t_i}-\frac{GV}{N},
\end{equation}
if MeanFlow is bounded $\|\bm{\mu}_{t_i}\|\le V, V \in \mathbb{R}_+$. Such boundedness assumptions are standard (Theorem 1.1 in \citep{gao2024convergence}, Lemma 7 in \citep{mena2025statistical}),  in theoretical analyses of flow-based generative models and hold, for example, for rectified-flow velocity fields on bounded domains, where the velocity norm is controlled by the domain diameter. In practice, the assumption can also be enforced by bounded network outputs or output clipping. With \cref{eq:weighting_schedule} and \cref{eq:ineq_h_rho}, the upper bound of scheduled deficit $d_i$ can be derived as
\begin{equation}
    d_i \le \frac{\bar{h}+GV}{N}
\end{equation}
With the non-zero gradient $\|\nabla h(\hat{\bm{\tau}}_{t_i})\| \ge m, m \in \mathbb{R}_+$, the minimum correction of $\bm{u}_{t_i}$ by \cref{eq:barrier_const_deficit} can be written as
\begin{equation}
    ||\bm{u}_{t_i}\Delta t||=\frac{d_i}{\|\nabla h(\hat{\bm{\tau}}_{t_i})\|} \le \frac{\bar{h}+GV}{mN}
\end{equation}
since $\Delta t=\frac{1}{N}$, we can finally obtain $\bar{U}$ as
\begin{equation}
    ||\bm{u}_{t_i}|| \le \bar{U}, 
    \quad
    \bar{U}=\frac{\bar{h}+GV}{m},
\end{equation} 
which can be estimated by the property of constraints $h(\cdot)$ and MeanFlow model $\bm{\mu}^{\bm{\theta}}_{t_i}$.

\section{Additional Evaluation} \label{append:other_baseline}
\subsection{Performance Across Navigation, Locomotion, and Manipulation}

We further evaluate ASAP against the following baselines across navigation, locomotion, and manipulation tasks:
\begin{itemize}
\item \textbf{Flow Matching (FM)}~\citep{lipman2022flow} generates trajectories directly with the pretrained flow model without safety enforcement.
\item \textbf{Truncation}~\cite{locomotion_env} modifies diffusion-generated trajectories to satisfy safety constraints.
\item \textbf{Rejection Sampling} samples a batch of candidate trajectories by a pretrained MeanFlow model and selects the safest one.
\end{itemize}

As shown in \Cref{tab:other_baseline_all_benchmark}, ASAP is the only method that satisfies safety across all benchmarks while maintaining competitive task performance. Flow matching sometimes achieves the highest score of the task, which is benefited by its multi-step sampling, but provides no safety guarantee. Truncation produces safe trajectories on some tasks but still lacks guarantees and requires multi-step sampling, inducing a higher computational cost. Rejection sampling can be combined with MeanFlow for efficient generation, yet cannot reliably satisfy unseen test-time constraints because safety depends on sufficiently safe candidates from the learned distribution. Overall, these results demonstrate that ASAP uniquely combines efficient planning with consistent safety across diverse robotic tasks.

\input{tables/append_other_baseline}

\subsection{Performance on High-Dimensional Tasks}\label{append:append_adroit}

We evaluate ASAP and all baselines on the Adroit Relocate task, a high-dimensional dexterous manipulation benchmark in which a robotic hand picks and relocates a ball. The task has a 39-dimensional state space and a 30-dimensional action space. As shown in \Cref{tab:append_adroit}, ASAP is the fastest method among all methods that guarantees safety. SAD-Flower and SafeDiffuser also satisfy the safety constraints and achieve higher task scores, but incur substantially greater computational cost due to repeated multi-step prediction and optimization. Other baselines violate the safety constraints and are also slower than ASAP because they require multiple sampling steps. MeanFlow is capable of planning faster marginally but cannot guarantee safety. These results demonstrate that ASAP maintains safety with low computational cost even in high-dimensional robotic tasks.

\input{tables/append_adroit_result}

\input{tables/append_real_world}

\subsection{Real-World Robotic Planning Results}\label{append:real_world}
We provide the complete evaluation of the real-world robotic planning experiment introduced in \cref{sec:exp}. The task requires a Franka robot to pick an ice cream from a plate and deliver it through a shop window to a target hand position while avoiding a juicer placed as an unseen obstacle. Demonstrations are collected by a teleoperation system without this test-time constraint, and deployment assumes that the ice-cream and target hand positions are provided by perception. We evaluate $20$ trials across two initial plate locations and three target hand positions. We assume a perception system, e.g., motion capture or a depth camera, provides accurate object states. In our deployment, we manually specify the ice cream and target hand locations.

We evaluate each method in terms of safety, planning time, and task performance. Safety is measured by the minimum collision margin along the generated trajectory, normalized to $[0,1]$. Task performance is determined by the final placement error, with a score of $1$ for delivery within $1.0$~cm of the target and higher scores for greater accuracy. We compare ASAP with MeanFlow (MF), Classifier Guidance (CG), and SafeDiffuser (S-Diff.), representing strong baselines in planning efficiency, task performance, and safety, respectively, and additionally with Diffuser (Diff.), SafeFlowMatcher (SFM), and SAD-Flower (SAD-F). We introduce an additional baseline, Rejection-Sampling (RS) based on MeanFlow, which samples a batch of trajectory candidates with MeanFlow, and selects the one that is closest to the safe set. RS can achieve fast planning due to a pretrained MeanFlow model, but cannot guarantee the safety since it fully relies on the dataset to include safety constraints.

The complete results are reported in \Cref{tab:add_real_world}. Across the evaluated configurations, ASAP provides the best overall combination of safety, planning efficiency, and task performance. MF achieves fast planning but cannot reliably satisfy the unseen obstacle constraint, while CG maintains strong task performance yet similarly lacks safety guarantees. Both of the S-Diff and SAD-Flower satisfy the safety constraint but require substantially longer planning time due to their multi-step safety corrections. SafeFlowMatcher relies on its time-varying parameter settings and can generate unsafe trajectories sometimes. In contrast, ASAP consistently maintains collision-free trajectories while preserving high task performance and low planning latency. These results further demonstrate that the arbitrary-step safety formulation transfers effectively from simulation to real-world manipulation under unseen constraints.

\input{tables/append_time_horizon}
\input{figures/append_conservatism}
\input{tables/append_time_constraint_num}

\subsection{Computational Scalability of the QP.}\label{append:comp_hor_con}
ASAP formulates a control-augmented MeanFlow and computes the control through a quadratic program (QP). To analyze the factors affecting QP computation time, we vary the trajectory horizon in Hopper, which increases the number of optimization variables, and the number of safety constraints in Maze2d. All evaluations use the same hardware and software settings as the main experiments in \Cref{sec:exp}. As shown in \Cref{tab:comp_horizon} and \Cref{tab:comp_constraint_num}, increasing either the trajectory horizon or the number of constraints increases the QP solving time only modestly. These results demonstrate that the computational cost of ASAP scales favorably with both trajectory dimensionality and constraint number, supporting its applicability to more complex planning tasks.

\subsection{Practical Evaluation of Conservatism}\label{append:conservative}
To evaluate the practical effect of the zero conservatism established in \Cref{th:zero_conservatism}, we examine whether ASAP avoids unnecessary clearance from the safety boundary across $N\in{1,2,3,4,5}$ in Maze2d. We quantify this using a geometric safety margin (sm.), defined as the mean distance to the safety boundary over the 10 trajectory points closest to it, where smaller values indicate less conservative clearance. As shown in \Cref{fig:append_conservatism}, the margin remains consistently small and nearly unchanged across $N$, indicating that few-step generation does not introduce excessive geometric conservatism. Meanwhile, the score varies only slightly, showing that task performance is also insensitive to the number of generation steps.

\subsection{Comparison with Different SDF-based Projection}\label{append:compare_sdf_projection}
To evaluate the benefit of intermediate safety corrections in our ASAP, we compare with two alternatives using the same pretrained MeanFlow model: (i) \textbf{Final-Only}, which applies SDF projection only after the final generation step, and (ii) \textbf{Full}, which enforces safety through SDF projection at every generation step without intermediate relaxation. We evaluate these methods across all benchmarks using the same sampling steps and safety constraints, as reported in \Cref{tab:append_cbf_projection}.

For one-step generation ($N=1$), all three methods are mathematically equivalent and achieve identical safety, task performance, and similar planning times. For multi-step generation ($N>1$), all methods maintain safety, while ASAP achieves the highest task scores across KUKA, Adroit, and real-world handover. In particular, ASAP improves the KUKA score to $1.75$, compared with $1.15$ for Final-Only and $1.45$ for Full, and achieves $1.00$ in real-world handover, compared with $0.80$ for Final-Only and $0.70$ for Full. These improvements suggest that ASAP's time-varying relaxation enables progressive safety guidance that influences subsequent MeanFlow predictions, potentially avoiding large final corrections without enforcing full safety at every intermediate step. Thus, ASAP provides a flexible safety framework for both efficient one-step and multi-step generation.

\input{tables/append_cbf_projection}

\input{tables/append_closed_form}

\subsection{Closed-form Implementation}
\label{append:closed_form}
To evaluate the computational overhead of the QP solver, we compare the original ASAP implementation using a general-purpose QP solver with a closed-form implementation (ASAP (CF)) derived from \cref{eq:minimum_correct_SDF}. For independent state-wise safety constraints, the minimum-norm correction admits an analytical solution, eliminating the need for numerical QP optimization. As shown in \Cref{tab:append_closed_form}, ASAP (CF) reduces planning time across all benchmarks, achieving approximately $2$--$7.5\times$ speedups while maintaining zero safety violations and comparable task performance. These results demonstrate that ASAP can be further accelerated through closed-form corrections, while the general QP formulation retains flexibility for handling multiple coupled safety constraints.

\subsection{Comparison of Few-Step and Multi-Step Safety Corrections}
\label{append:few_step_comparison}
To investigate whether existing safety-guaranteed planners can directly benefit from few-step generation, we evaluate SAD-Flower (SAD-F), SafeFlowMatcher (SFM), and SafeDiffuser (S-Diff.) using few-step (F) settings matched to ASAP and their default multi-step (M) settings from the original papers and released code. The complete results across all benchmarks are reported in \Cref{tab:append_few_step_safe}.

ASAP achieves zero safety violations across all benchmarks with competitive task performance and low planning costs. In contrast, SAD-Flower exhibits safety violations under few-step generation, as its continuous-time safety formulation relies on repeated corrections. Similarly, SFM benefits from a single MeanFlow prediction but generally requires additional safety corrections to maintain safety, introducing substantial computational overhead. SafeDiffuser maintains safety in several few-step settings, but its aggressive corrections substantially degrade task performance, particularly on Hopper, KUKA, and real-world handover. Increasing the correction steps generally improves safety and task performance, but significantly increases planning time.

These results demonstrate that simply combining MeanFlow with existing safety mechanisms does not preserve both safety and task performance under few-step generation. In contrast, ASAP's discrete-time barrier formulation enables efficient safety corrections with formal guarantees for arbitrary sampling steps, including one-step generation.

\input{tables/append_safe_multi_step}

\input{tables/append_env_setting}
\section{Environment Details}\label{append:env} 
We evaluate ASAP across diverse trajectory-planning domains, summarized in \cref{tab:env_setting}: navigation (Maze2d-Large), locomotion (Hopper), robotic manipulation (KUKA Block-Stacking), and dexterous manipulation (Adroit Relocate). The environments are simulated in MuJoCo \citep{todorov2012mujoco} or PyBullet \citep{coumans2016pybullet} and span different state and action dimensions. For Maze2d, Hopper, and KUKA, we follow the default planning horizon and sampling steps in \citep{xiao2023safediffuser}. For Adroit, we use a shorter horizon and the same number of sampling steps as Hopper, since both jointly plan states and actions.

We train the generative and control models using offline datasets summarized in \cref{tab:dataset_setting}. The navigation, locomotion, and Adroit datasets are obtained from D4RL \citep{maze_env}, with dataset variants such as medium and medium-expert. Adroit Relocate data are generated by a fine-tuned RL policy \citep{rajeswaran2017learning}, while Hopper uses soft actor-critic (SAC) trajectories \citep{haarnoja2018soft} mixed with expert demonstrations. For KUKA Block-Stacking \citep{janner2022planning}, trajectories are generated using PDDLStream \citep{garrett2020pddlstream}, which provides feasible manipulation plans for structured stacking tasks.

\input{tables/append_env_dataset}

\section{Test-Time Safety Constraints}

We evaluate all methods under safety constraints that are introduced only at test time and are not represented in the training datasets.

\paragraph{Maze2d-Large.}
The planner generates a trajectory from a randomly sampled initial state to a randomly sampled goal while navigating the maze. To evaluate unseen constraint handling, we introduce two circular obstacles at test time that restrict the feasible region without fully blocking the path. Each obstacle is represented by the SDF
\begin{equation}
h(x,y)=\sqrt{(x-c_x)^2+(y-c_y)^2}-r,
\end{equation}
where $(x,y)\in\mathbb{R}^2$ is the planar state, $(c_x,c_y)$ is the obstacle center, and $r>0$ is its radius. Safety requires $h(x,y)\geq0$.

\paragraph{Hopper.}
For locomotion, we impose an unseen torso-height constraint to emulate overhead obstacles. Let $s$ denote the torso height. Safety requires
\begin{equation}
s \leq z,
\end{equation}
with $z=1.6$. This constraint conflicts with the forward-locomotion objective, making Hopper particularly challenging because satisfying safety can reduce task performance.

\paragraph{KUKA Block-Stacking.}
For KUKA manipulation, the generative model is conditioned on object locations and predicts joint-space trajectories. We impose tighter joint limits at test time to emulate workspace restrictions or partial system degradation:
\begin{equation}
\bm{q}\in\beta[\bm{q}_{\min},\bm{q}_{\max}],
\end{equation}
where $\beta\in\{0.9,0.8,0.7\}$ controls the constraint severity, which maps to the constraint level $=\{10\%,20\%,30\%\}$ in the \Cref{fig:constraint_levels}. Smaller $\beta$ values progressively restrict the feasible joint range and allow us to evaluate robustness under increasingly strict unseen constraints.

\paragraph{Adroit Relocate.}
For dexterous manipulation, the policy is conditioned on the initial object position and relocation target and jointly plans state-action trajectories. To emulate an unsafe workspace region, we impose the unseen Cartesian constraint
\begin{equation}
x \leq 0.1,
\end{equation}
which restricts the arm workspace during relocation.

Together, these constraints evaluate whether each method can satisfy previously unseen safety requirements while preserving task performance across navigation, locomotion, and manipulation.

\section{Training Details} \label{append:train}

We provide the implementation and training details for all evaluated methods, including codebases, optimization settings, and model architectures.

\paragraph{Implementations.}
We implement each baseline using its official or publicly available codebase whenever possible. Diffuser follows \citet{janner2022planning}; Classifier Guidance and SafeDiffuser follow\citet{xiao2023safediffuser}; SAD-Flower follows \citet{huang2026sad}; SafeFlowMatcher follows \citet{yang2026safeflowmatcher} and Flow Matching follows \citet{feng2025on}. We implement CoBL with the algorithm described in \citet{mizuta2024cobl} and the rejection sampling method based on the pretrained MeanFlow model. For our method, we consider the codebase from Flow matching~\citet{feng2025on}, SAD-Flower~\citet{huang2026sad}, MeanFlow~\citet{geng2025mean}, and the MeanFlow policy~\citet{sheng2026mp1} and follow a similar framework as \citet{feng2025on}, enabling a direct comparison under a shared generative backbone.

\paragraph{Training Configuration.}
The training hyperparameters used for each environment are summarized in \Cref{tab:train_detail}. Within each environment, all methods use the same planning horizon. Our method uses the similar amount of training step as other baseline, so the comparison isolates the effect of our proposed safety mechanism rather than differences in model training.

\paragraph{Model Architecture.}
The diffusion-based baselines, including Diffuser, Classifier Guidance, CoBL, and SafeDiffuser, adopt the temporal U-Net architecture of \citet{janner2022planning}, with residual temporal convolutions, group normalization, and Mish activations. The SafeFlowMatcher also applies a temporal U-Net architecture following its source code. For flow-based baselines, including Flow Matching, SAD-Flower, and our method use the Transformer backbone of \citet{feng2025on}, with a hidden dimension of $256$, eight Transformer layers, eight attention heads, and an MLP intermediate dimension of $1024$.

\input{tables/append_train_detail}

\section{Computational Resources} \label{append:compute}
All simulation training and evaluation runs were conducted on a single platform with an AMD EPYC processor and an NVIDIA RTX 4090 GPU with 24 GB of memory. Training for the real-world robot-handing task used the same platform, while evaluation was performed on a separate local machine. A complete summary of the experimental hardware is provided in \cref{tab:append_comp}.

Across Maze2d, locomotion, KUKA, Adroit, and the real-world robot-handing task, the safety-guaranteed baselines, including SafeFlowMatcher~\citep{yang2026safeflowmatcher}, SAD-Flower~\citep{huang2026sad}, and SafeDiffuser~\citep{xiao2023safediffuser}, require QP optimization as ASAP does. For a fair planning-time comparison, all methods use the same QP solver, \emph{qpth}~\citep{amos2017optnet}. Because \emph{qpth} supports only QP formulations, the result of post-optimization ablation in \Cref{tab:qp_vs_post}, which involves a non-QP problem, instead evaluates both ASAP and the post-optimization baseline using the package in \emph{SciPy}. This package can be applied to ASAP to further improve the inference speed, which is validated in real-world experiments.  \looseness=-1
\input{tables/append_comp_resource}

%% file: tables/append_other_baseline.tex
\begin{table}[htbp]
\centering
\caption{\textbf{Comparison with other baselines across navigation, locomotion, and manipulation.} Safety violation (s-vio), planning time (s) and task score of Flow matching (FM), Truncation, Reject sampling and ASAP on Maze2d, Hopper and KUKA block stacking (the $30\%$ constraint level).}
\label{tab:other_baseline_all_benchmark}

\resizebox{0.9\textwidth}{!}{%
{\small
\begin{tabular}{
    >{\centering\arraybackslash}m{3.0cm}
    >{\centering\arraybackslash}m{1.2cm}
    >{\centering\arraybackslash}m{2.0cm}
    >{\centering\arraybackslash}m{2.0cm}
    >{\centering\arraybackslash}m{2.0cm}
    >{\centering\arraybackslash}m{2.0cm}
}
\toprule
\textbf{Experiment}  & \textbf{Metric} & \textbf{FM} &\textbf{Truncation} &\textbf{Reject sampling} & \textbf{ASAP} \\
\midrule
\multirow{4}{3cm}{\centering Maze2d (Large)} 
& s-vio ($\downarrow$)        & $0.41\pm0.34$  & $0.07\pm0.15$  & $0.15\pm0.23$                  & $\textbf{0.00}\bm{\pm}\textbf{0.00}$ \\
\cmidrule{2-6}
& time ($\downarrow$)     & $2.86\pm0.04$       & $5.43\pm2.94$  & $\textbf{0.01}\bm{\pm}\textbf{0.00}$             & $0.06\pm0.01$ \\
\cmidrule{2-6}
& score  ($\uparrow$)       & $\textbf{1.61}\bm{\pm}\textbf{0.01}$ & $1.50\pm0.25$ & $1.37\pm0.51$ & $1.46\pm0.47$ \\
\midrule
\multirow{4}{3cm}{\centering Hopper (ME)} 
& s-vio ($\downarrow$)        & $\textbf{0.00}\bm{\pm}\textbf{0.00}$    & $\textbf{0.00}\bm{\pm}\textbf{0.00}$ & $\textbf{0.00}\bm{\pm}\textbf{0.00}$                & $\textbf{0.00}\bm{\pm}\textbf{0.00}$ \\
\cmidrule{2-6}
& time  ($\downarrow$)    & $0.12\pm0.00$      & $0.15\pm0.00$ & $0.04\pm0.00$              & $\textbf{0.05}\bm{\pm}\textbf{0.00}$ \\
\cmidrule{2-6}
& score ($\uparrow$)       & $\textbf{0.44}\bm{\pm}\textbf{0.05}$ & $0.42\pm0.08$  & $0.43\pm0.12$ & $0.41\pm0.14$ \\
\midrule
\multirow{4}{3cm}{\centering KUKA Block Stacking ($30\%$ constraint)} 
& s-vio  ($\downarrow$)       & $0.92\pm0.17$  & $\textbf{0.00}\bm{\pm}\textbf{0.00}$ & $0.25\pm0.06$                & $\textbf{0.00}\bm{\pm}\textbf{0.00}$ \\
\cmidrule{2-6}
& time  ($\downarrow$)       & $9.70\pm0.15$ & $21.83\pm0.83$ & $0.06\pm0.00$  & $\textbf{0.10}\bm{\pm}\textbf{0.00}$ \\
\cmidrule{2-6}
& score  ($\uparrow$)       & $1.05\pm0.86$ & $\textbf{1.25}\bm{\pm}\textbf{1.09}$ & $1.10\pm0.83$  & $1.10\pm0.89$ \\

\bottomrule
\end{tabular}
}%
}
\end{table}

%% file: tables/append_adroit_result.tex
\begin{table}[!t]
\caption{\textbf{Adroit Relocate evaluation for all baselines.} Safety violation, planning time (s), and task score on dexterous manipulation with a 39-dimensional state space and a 30-dimensional action space.}
\label{tab:append_adroit}
\vspace{-0.3em}
\resizebox{\textwidth}{!}{ 
\begin{tabular}{
    >{\centering\arraybackslash}m{1.5cm} 
    >{\centering\arraybackslash}m{1.3cm} 
    >{\centering\arraybackslash}m{1.4cm} 
    >{\centering\arraybackslash}m{1.4cm} 
    >{\centering\arraybackslash}m{1.4cm} 
    >{\centering\arraybackslash}m{1.4cm} 
    >{\centering\arraybackslash}m{1.4cm} 
    >{\centering\arraybackslash}m{1.4cm} 
    >{\centering\arraybackslash}m{1.4cm} 
    >{\centering\arraybackslash}m{1.4cm} 
}
\toprule
\textbf{Task}  & \textbf{Metric} & \textbf{MF} & \textbf{SFM} & \textbf{SAD-F}  & \textbf{Diffuser} & \textbf{CG} & \textbf{S-Diff.} & \textbf{CoBL} & \textbf{ASAP} \\
\midrule
\multirow{4}{2cm}{\centering Adroit (Relocate)} 
& s-vio ($\downarrow$)  &0.36$\pm$0.07 & 0.12$\pm$0.00 & \textbf{0.00$\pm$0.00} &0.01$\pm$0.02&0.12$\pm$0.00& \textbf{0.00}$\bm{\pm}$\textbf{0.00} & 0.02$\pm$0.03& \textbf{0.00}$\bm{\pm}$\textbf{0.00} \\
\cmidrule{2-10}
& time ($\downarrow$) & \textbf{0.02}$\bm{\pm}$\textbf{0.00} & 0.28$\pm$0.00 & 1.54$\pm$0.15 &0.13$\pm$0.00& 0.14$\pm$0.00 &0.59$\pm$0.01& 0.16$\pm$0.01 & 0.15$\pm$0.00 \\
\cmidrule{2-10}
& score ($\uparrow$) &1.07$\pm$0.12 &0.18$\pm$0.26 &1.05$\pm$0.23& 0.91$\pm$0.41& 0.78$\pm $0.39&1.07$\pm$0.08 & 0.01$\pm$0.02& \textbf{1.10}$\bm{\pm}$\textbf{0.06}\\
\bottomrule\bottomrule
\end{tabular}
}
\end{table}

%% file: tables/append_real_world.tex
\begin{table}[b] 
\centering
\caption{\textbf{Real-world handover for all baselines.} Safety violation (s-vio), planning time (s), and task score over $20$ trials.}
\resizebox{1\linewidth}{!}{
\label{tab:add_real_world}
\begin{tabular}{
    >{\centering\arraybackslash}m{1.2cm} 
    >{\centering\arraybackslash}m{1.3cm} 
    >{\centering\arraybackslash}m{1.4cm}
    >{\centering\arraybackslash}m{1.4cm}
    >{\centering\arraybackslash}m{1.4cm}
    >{\centering\arraybackslash}m{1.4cm}
    >{\centering\arraybackslash}m{1.4cm} 
    >{\centering\arraybackslash}m{1.4cm} 
    >{\centering\arraybackslash}m{1.4cm}
    >{\centering\arraybackslash}m{1.4cm}
}
\toprule
\textbf{Task}  & \textbf{Metric} & \textbf{MF} & \textbf{SFM} & \textbf{SAD-F} & \textbf{CG} & \textbf{RS}& \textbf{Diff.} & \textbf{S-Diff.} & \textbf{ASAP} \\
\midrule
\multirow{4}{2cm}{Handing \\ice cream} 
& s-vio ($\downarrow$) &0.35$\pm$0.11 &0.01$\pm$0.00 &\textbf{0.00}$\bm{\pm}$\textbf{0.00} & 0.27$\pm$0.12 & 0.28$\pm$0.08 & 0.27$\pm$0.13 &\textbf{0.00}$\bm{\pm}$\textbf{0.00} & \textbf{0.00}$\bm{\pm}$\textbf{0.00} \\
\cmidrule{2-10}
& time ($\downarrow$) & \textbf{0.01}$\bm{\pm}$\textbf{0.01} & 2.31$\pm$0.26 & 2.28$\pm$0.27 & 2.18$\pm$0.23 & \textbf{0.01}$\bm{\pm}$\textbf{0.01} &0.86$\pm$0.08 & 13.23$\pm$1.35 & 0.04$\pm$0.01 \\
\cmidrule{2-10}
& score ($\uparrow$) & 0.30$\bm{\pm}$0.03 & 0.74$\pm$0.43& \textbf{1.00}$\bm{\pm}$\textbf{0.01} &0.30$\pm$0.02 & 0.28$\pm$0.10& 0.28$\pm$0.07 & 0.94$\pm$0.22 & \textbf{1.00}$\bm{\pm}$\textbf{0.01}\\
\bottomrule\bottomrule
\end{tabular}
}

\end{table}

%% file: tables/append_time_horizon.tex
\begin{table}[t]
\centering
\caption{\textbf{The computational cost of ASAP with varying horizon length on Hopper.} \looseness=-1}
\resizebox{0.9\textwidth}{!}{
\label{tab:comp_horizon}
\begin{tabular}{lccccc}
\toprule
 & $H=20$ & $H=40$ & $H=60$ & $H=80$ & $H=100$ \\
\midrule
time (s)     & $0.034\pm0.001$             & $0.037\pm0.000$                 & $0.040\pm0.001$    & $0.041\pm0.000$ & $0.042\pm0.000$             \\
\bottomrule
\end{tabular}
}
\end{table}

%% file: figures/append_conservatism.tex
\begin{figure}[b]\vspace{0pt}
        \captionsetup{justification=raggedright, singlelinecheck=false}%
        \centering
        {\figfont{7}\pgfplotslegendfromname{asapleft}}\par
        \vspace{1pt}
        \begin{tikzpicture}
            \begin{axis}[
                asap,
                name=leftaxis,
                width=0.57\linewidth, height=47pt,
                xlabel={Sampling steps N},
                xlabel style={yshift=2pt, font=\figfont{8}},
                xmin=0.5, xmax=5.5, xtick={1,2,3,4,5},
                ymin=0, ymax=0.010, ytick={0,0.005,0.010},
                yticklabel style={/pgf/number format/fixed,
                                  /pgf/number format/precision=3,
                                  font=\figfont{7}, text=cons_color},
                xticklabel style={font=\figfont{7}},
                scaled y ticks=false,
                legend to name=asapleft, legend columns=2,
                legend style={draw=none, fill=none, font=\figfont{7},
                              at={(0,0)}, anchor=center, /tikz/row sep=-1pt,
                              /tikz/every odd column/.append style={column sep=0.15em},
                              /tikz/every even column/.append style={column sep=0.9em}},
            ]
                \addplot[asap ours, color=cons_color, mark options={solid, fill=cons_color, draw=fig_edge, line width=0.45pt}] table[x=N_step, y=conservatism]
                    {figures/data/left_tradeoff_points.txt};
                \addlegendentry{Sm.}
                \addlegendimage{color=score_color, line width=0.9pt, mark=square*,
                    mark size=1.8pt, mark options={solid, fill=white,
                    draw=score_color, line width=0.7pt}}
                \addlegendentry{Score}
            \end{axis}
            \begin{axis}[
                asap,
                name=rightaxis,
                width=0.57\linewidth, height=47pt,
                axis y line*=right, axis x line=none, ymajorgrids=false,
                xmin=0.5, xmax=5.5,
                ymin=1.40, ymax=1.50, ytick={1.40,1.45,1.50},
                yticklabel style={/pgf/number format/fixed,
                                  /pgf/number format/precision=2,
                                  font=\figfont{7}, text=score_color},
                scaled y ticks=false,
                y axis line style={line width=0.6pt, draw=fig_ink!85},
            ]
                \addplot[color=score_color, line width=0.9pt, mark=square*,
                         mark size=1.8pt, mark options={solid, fill=white,
                         draw=score_color, line width=0.7pt}]
                    table[x=N_step, y=score] {figures/data/left_tradeoff_points.txt};
            \end{axis}
        \end{tikzpicture}\par
    \caption{\textbf{Safety margin (sm.) across sampling steps in Maze2d.} left axis: geometry safety margin to the obstacle boundary;
        blue, right axis: task score. Both flat from $N=1$ to $5$.}
    \label{fig:append_conservatism}
\end{figure}
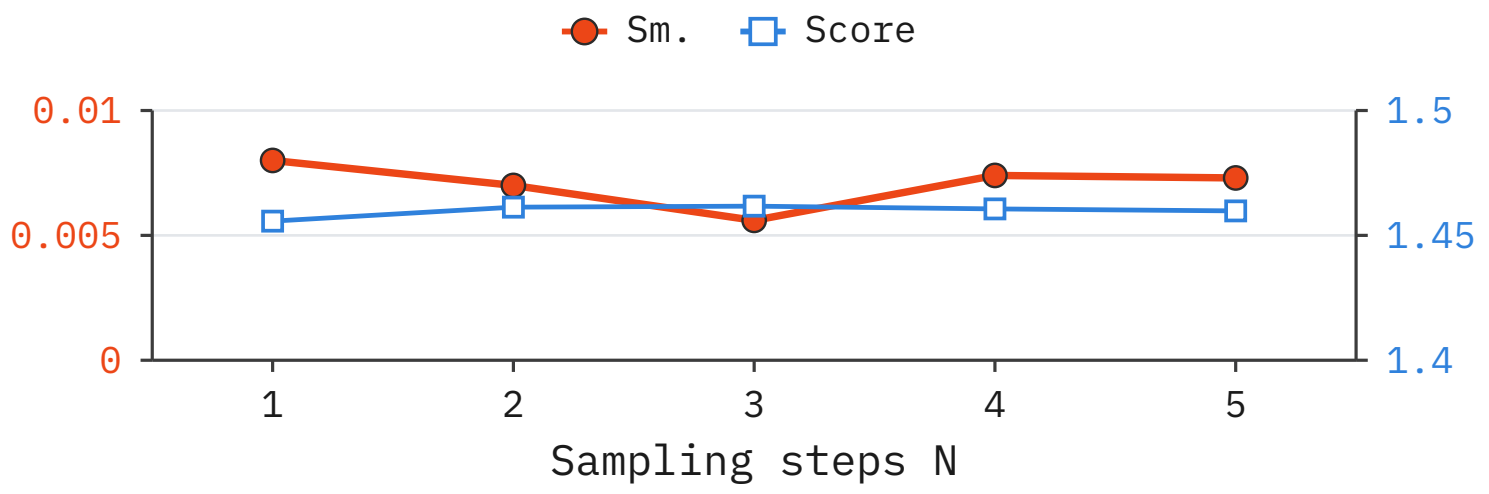

%% file: tables/append_time_constraint_num.tex
\begin{table}[t]
\centering
\caption{\textbf{The computational cost of ASAP with varying amounts of constraints $N_C$ on Maze2d.} \looseness=-1}
\resizebox{0.6\textwidth}{!}{
\label{tab:comp_constraint_num}
\begin{tabular}{lccc}
\toprule
 & $N_C=1$ & $N_C=2$ & $N_C=4$ \\
\midrule
time (s)     & $0.056\pm0.002$             & $0.056\pm0.007$    & $0.063\pm0.003$             \\
\bottomrule
\end{tabular}
}
\end{table}

%% file: tables/append_cbf_projection.tex
\begin{table}[t]
\centering
\caption{\textbf{Comparison with SDF-based projection baselines across all benchmarks.} Safety violation (s-vio), planning time (s) and task score of MeanFlow followed by final-step only SDF projection (Final-Only), MeanFlow with full SDF projection at every step (Full) and ASAP on Maze2d, Hopper, KUKA block stacking (the $10\%$ constraint level), Adroit, and Real-world task for handing ice cream via a Franka robot.}
\label{tab:append_cbf_projection}
\resizebox{0.95\textwidth}{!}{%
{\small
\begin{tabular}{
    >{\centering\arraybackslash}m{4.0cm}
    >{\centering\arraybackslash}m{1.2cm}
    >{\centering\arraybackslash}m{2.0cm}
    >{\centering\arraybackslash}m{2.0cm}
    >{\centering\arraybackslash}m{2.0cm}
}
\toprule
\textbf{Experiment}  & \textbf{Metric} & \textbf{Final-Only} &\textbf{Full} & \textbf{ASAP} \\
\midrule
\multirow{4}{3cm}{\centering Maze2d (Large) \\ Sampling Step$=$1} 
& s-vio ($\downarrow$)       & $0.00\pm0.00$ & $0.00\pm0.00$                  & $0.00\pm0.00$ \\
\cmidrule{2-5}
& time ($\downarrow$)           & $0.06\pm0.01$  & $0.06\pm0.00$             & $0.06\pm0.01$ \\
\cmidrule{2-5}
& score  ($\uparrow$)      & $1.46\pm0.47$ & $1.46\pm0.47$ & $1.46\pm0.47$ \\
\midrule
\multirow{4}{3cm}{\centering Hopper (ME)\\ Sampling Step$=$1} 
& s-vio ($\downarrow$)     & $0.00\pm0.00$ & $0.00\pm0.00$     & $0.00\pm0.00$ \\
\cmidrule{2-5}
& time  ($\downarrow$)    & $0.05\pm0.00$       & $0.05\pm0.01$  & $0.05\pm0.00$ \\ 
\cmidrule{2-5}
& score ($\uparrow$)       & $0.41\pm0.14$ & $0.41\pm0.14$   & $0.41\pm0.14$ \\
\midrule
\multirow{4}{3cm}{\centering KUKA Block Stacking ($10\%$ constraint)\\ Sampling Step$=$4} 
& s-vio  ($\downarrow$)       & $0.00\pm0.00$ & $0.00\pm0.00$     & $0.00\pm0.00$ \\
\cmidrule{2-5}
& time  ($\downarrow$)       & $0.05\pm0.00$ & $0.10\pm0.00$ & $0.10\pm0.00$  \\
\cmidrule{2-5}
& score  ($\uparrow$)       & $1.15\pm0.79$ & $1.45\pm0.92$ & $1.75\pm0.83$  \\
\midrule
\multirow{4}{3cm}{\centering Adroit\\ Sampling Step$=$3} 
& s-vio ($\downarrow$)     & $0.00\pm0.00$ & $0.00\pm0.00$     & $0.00\pm0.00$ \\
\cmidrule{2-5}
& time  ($\downarrow$)    & $0.06\pm0.00$       & $0.15\pm0.00$  & $0.15\pm0.00$ \\ 
\cmidrule{2-5}
& score ($\uparrow$)       & $1.06\pm0.12$ & $1.08\pm0.07$   & $1.10\pm0.06$ \\
\midrule
\multirow{4}{3cm}{\centering Real-world robot handover\\ Sampling Step$=$2} 
& s-vio ($\downarrow$)     & $0.00\pm0.00$ & $0.00\pm0.00$     & $0.00\pm0.00$ \\
\cmidrule{2-5}
& time  ($\downarrow$)    & $0.02\pm0.00$       & $0.04\pm0.00$  & $0.04\pm0.01$ \\ 
\cmidrule{2-5}
& score ($\uparrow$)       & $0.80\pm0.40$ & $0.70\pm0.46$   & $1.00\pm0.01$ \\
\midrule
\bottomrule
\end{tabular}
}
}%
\end{table}

%% file: tables/append_closed_form.tex
\begin{table}[t]
\centering
\caption{\textbf{Comparison with implementations of closed-form and optimization across all benchmarks.} Safety violation (s-vio), planning time (s) and task score of MeanFlow followed by closed-form implementation of ASAP (\textbf{ASAP (CF)}) and QP solver implementation (\textbf{ASAP}) on Maze2d, Hopper, KUKA block stacking (the $10\%$ constraint level), Adroit, and a real-world task for handing ice cream via Franka robot.}
\label{tab:append_closed_form}
\resizebox{0.75\textwidth}{!}{%
{\small
\begin{tabular}{
    >{\centering\arraybackslash}m{3.0cm}
    >{\centering\arraybackslash}m{1.2cm}
    >{\centering\arraybackslash}m{2.0cm}
    >{\centering\arraybackslash}m{2.0cm}
}
\toprule
\textbf{Experiment}  & \textbf{Metric} &\textbf{ASAP (CF)} & \textbf{ASAP} \\
\midrule
\multirow{4}{3cm}{\centering Maze2d (Large)} 
& s-vio ($\downarrow$)       & $0.00\pm0.00$                  & $0.00\pm0.00$ \\
\cmidrule{2-4}
& time ($\downarrow$)          & $0.01\pm0.00$             & $0.06\pm0.01$ \\
\cmidrule{2-4}
& score  ($\uparrow$)      & $1.46\pm0.47$ & $1.46\pm0.47$ \\
\midrule
\multirow{4}{3cm}{\centering Hopper (ME)} 
& s-vio ($\downarrow$)    & $0.00\pm0.00$     & $0.00\pm0.00$ \\
\cmidrule{2-4}
& time  ($\downarrow$)        & $0.01\pm0.00$  & $0.05\pm0.00$ \\ 
\cmidrule{2-4}
& score ($\uparrow$)        & $0.41\pm0.14$   & $0.41\pm0.14$ \\
\midrule
\multirow{4}{3cm}{\centering KUKA Block Stacking ($10\%$ constraint)} 
& s-vio  ($\downarrow$)      & $0.00\pm0.00$     & $0.00\pm0.00$ \\
\cmidrule{2-4}
& time  ($\downarrow$)       & $0.03\pm0.00$ & $0.10\pm0.00$  \\
\cmidrule{2-4}
& score  ($\uparrow$)      & $1.75\pm0.89$ & $1.75\pm0.83$  \\
\midrule
\multirow{4}{3cm}{\centering Adroit} 
& s-vio ($\downarrow$)     & $0.00\pm0.00$     & $0.00\pm0.00$ \\
\cmidrule{2-4}
& time  ($\downarrow$)         & $0.02\pm0.00$  & $0.15\pm0.00$ \\ 
\cmidrule{2-4}
& score ($\uparrow$)      & $1.07\pm0.10$   & $1.10\pm0.06$ \\
\midrule
\multirow{4}{3cm}{\centering Real-world robot handover} 
& s-vio ($\downarrow$)     & $0.00\pm0.00$     & $0.00\pm0.00$ \\
\cmidrule{2-4}
& time  ($\downarrow$)         & $0.02\pm0.00$  & $0.04\pm0.01$ \\ 
\cmidrule{2-4}
& score ($\uparrow$)       & $1.00\pm0.01$   & $1.00\pm0.01$ \\
\midrule
\bottomrule
\end{tabular}
}
}%
\end{table}

%% file: tables/append_safe_multi_step.tex
\begin{table}[t]
\centering
\caption{\textbf{Comparison of few-step and multi-step safety corrections across all benchmarks.} The baselines SAD-F, SFM, and S-Diff. are evaluated with both few-step (F) as ASAP and multi-step (M) safety corrections using the default settings from the original papers and released code.}
\label{tab:append_few_step_safe}
\resizebox{1\textwidth}{!}{%
{\small
\begin{tabular}{
    >{\centering\arraybackslash}m{3.0cm}
    >{\centering\arraybackslash}m{1.2cm}
    >{\centering\arraybackslash}m{1.5cm}
    >{\centering\arraybackslash}m{1.7cm}
    >{\centering\arraybackslash}m{1.5cm}
    >{\centering\arraybackslash}m{1.6cm}
    >{\centering\arraybackslash}m{1.5cm}
    >{\centering\arraybackslash}m{1.6cm}
    >{\centering\arraybackslash}m{1.5cm}
}
\toprule
\textbf{Experiment}  & \textbf{Metric} &\textbf{SAD-F (F)}&\textbf{SAD-F (M)}&\textbf{SFM (F)}&\textbf{SFM (M)} &\textbf{S-Diff. (F)} &\textbf{S-Diff. (M)} & \textbf{ASAP (F)} \\
\midrule
\multirow{4}{3cm}{\centering Maze2d (Large)} 
& s-vio ($\downarrow$)                         & $0.57\pm0.36$ & $0.00\pm0.00$& $0.05\pm0.08$& $0.71\pm0.20$& $0.00\pm0.00$& $0.00\pm0.00$ & $0.00\pm0.00$\\
\cmidrule{2-9}
& time ($\downarrow$)          & $0.10\pm0.01$ & $35.84\pm2.51$& $0.11\pm0.00$& $15.50\pm0.75$& $0.86\pm0.18$& $17.08\pm2.29$             & $0.06\pm0.01$ \\
\cmidrule{2-9}
& score  ($\uparrow$)      & $1.61\pm0.07$& $1.44\pm0.28$& $0.76\pm0.66$& $1.30\pm0.67$& $0.80\pm0.57$& $1.46\pm0.23$ & $1.46\pm0.47$ \\
\midrule
\multirow{4}{3cm}{\centering Hopper (ME)} 
& s-vio ($\downarrow$)    & $0.05\pm0.03$ & $0.00\pm0.00$& $0.07\pm0.04$& $0.04\pm0.02$& $0.00\pm0.00$& $0.00\pm0.00$     & $0.00\pm0.00$ \\
\cmidrule{2-9}
& time  ($\downarrow$)        & $0.05\pm0.01$  & $1.66\pm0.03$  & $0.02\pm0.00$  & $1.13\pm0.01$  & $0.05\pm0.01$  & $17.08\pm2.29$  & $0.05\pm0.00$ \\ 
\cmidrule{2-9}
& score ($\uparrow$)        & $0.41\pm0.14$  & $0.43\pm0.11$  & $0.33\pm0.15$  & $0.39\pm0.09$  & $0.13\pm0.05$  & $0.44\pm0.07$   & $0.41\pm0.14$ \\
\midrule
\multirow{4}{3cm}{\centering KUKA Block Stacking ($10\%$ constraint)} 
& s-vio  ($\downarrow$)      & $0.04\pm0.08$  & $0.00\pm0.00$ & $0.00\pm0.00$ & $0.00\pm0.00$ & $0.00\pm0.00$ & $0.00\pm0.00$     & $0.00\pm0.00$ \\
\cmidrule{2-9}
& time  ($\downarrow$)       & $0.02\pm0.00$ & $37.21\pm0.08$& $0.09\pm0.00$& $14.61\pm0.49$& $0.15\pm0.00$& $68.75\pm1.65$ & $0.10\pm0.00$  \\
\cmidrule{2-9}
& score  ($\uparrow$)      & $1.15\pm0.65$ & $1.45\pm0.67$& $1.45\pm0.60$& $1.10\pm0.70$& $0.55\pm0.67$& $1.60\pm0.86$ & $1.75\pm0.83$  \\
\midrule
\multirow{4}{3cm}{\centering Adroit} 
& s-vio ($\downarrow$)     & $0.05\pm0.03$ & $0.07\pm0.04$ & $0.12\pm0.00$ & $0.00\pm0.00$ & $0.00\pm0.00$ & $0.00\pm0.00$     & $0.00\pm0.00$ \\
\cmidrule{2-9}
& time  ($\downarrow$)         & $0.05\pm0.00$  & $1.54\pm0.15$  & $0.02\pm0.00$  & $0.28\pm0.00$  & $0.05\pm0.00$  & $0.59\pm0.01$  & $0.15\pm0.00$ \\ 
\cmidrule{2-9}
& score ($\uparrow$)      & $1.07\pm0.10$ & $1.07\pm0.10$ & $1.07\pm0.10$ & $1.07\pm0.10$ & $1.07\pm0.10$ & $1.07\pm0.10$   & $1.08\pm0.07$ \\
\midrule
\multirow{4}{3cm}{\centering Real-world robot handover} 
& s-vio ($\downarrow$)     & $0.04\pm0.03$ & $0.00\pm0.00$& $0.36\pm0.21$& $0.45\pm0.25$& $0.00\pm0.00$& $0.00\pm0.00$     & $0.00\pm0.00$ \\
\cmidrule{2-9}
& time  ($\downarrow$)         & $0.10\pm0.14$ & $2.28\pm0.27$ & $0.06\pm0.01$ & $0.49\pm0.06$ & $0.14\pm0.02$ & $13.23\pm1.35$  & $0.04\pm0.01$ \\ 
\cmidrule{2-9}
& score ($\uparrow$)       & $0.00\pm0.00$ & $1.00\pm0.01$& $0.05\pm0.19$& $0.35\pm0.48$& $0.00\pm0.00$& $0.94\pm0.22$   & $1.00\pm0.01$ \\
\midrule
\bottomrule
\end{tabular}
}
}%
\end{table}

%% file: tables/append_env_setting.tex
\begin{table}[t]
\centering
\caption{\textbf{Benchmark settings.}}
\resizebox{1\textwidth}{!}{
\label{tab:env_setting}
\begin{tabular}{lcccccc}
\toprule
\textbf{Environment} & \textbf{Evaluation trials} & \textbf{Obs. Dim.} & \textbf{Action Dim.} & \textbf{Horizon} & \textbf{Default Sampling Steps} & \textbf{ASAP Sampling Steps} \\
\midrule
Maze2d-Large-v1     & 10             & 4                  & 2    & 384 & 256 & 1              \\
Hopper (Medium-Expert)             & 10             & 11                 & 3  & 600 & 20   & 1               \\
KUKA Block-Stacking & 20           & 39                  & -   & 128 &   1000      & 4         \\
Adroit Relocate (Expert) & 20           & 39                  & 30    & 32 & 20      & 3          \\
Real-world handover & 20 & 8 & 8 & 128 & 256 & 2 \\
\bottomrule
\end{tabular}
}
\end{table}

%% file: tables/append_env_dataset.tex
\begin{table}[t]
\centering
\caption{\textbf{Datasets.} Size and source or generation method of the training data for each benchmark.}
\resizebox{1\textwidth}{!}{
\label{tab:dataset_setting}
\begin{tabular}{lcc}
\toprule
\textbf{Environment} & \textbf{\# of Trajectories} & \textbf{Source / Generation Method} \\
\midrule
Maze2d-Large-v1 & $4 \times 10^6$ & D4RL \citep{maze_env} \\
Hopper (Medium-Expert) & $2 \times 10^6$ & Mixture of expert and partial SAC \citep{haarnoja2018soft} \\
KUKA Block-Stacking & 10,000 & PDDLStream planner \citep{garrett2020pddlstream} \\
Adroit Relocate (Expert) & $1 \times 10^6$ & Finetuned RL policy~\citep{rajeswaran2017learning} \\
\bottomrule
\end{tabular}
}
\end{table}

%% file: tables/append_train_detail.tex
\begin{table}[t]
\centering
\small
\caption{\textbf{Training configurations.} Batch size, learning rate and training steps of every method on each benchmark.}
\label{tab:train_detail}
\begin{tabular}{lcccccccc}
\toprule
\textbf{Maze2d-Large} & Diffuser  & CG & S-Diff. & FM & SAD-F & SFM & CoBL & ASAP \\
\midrule
Batch Size & 128  & 128 & 128 & 128 & 128 & 128 & 128 & 128 \\
Learning Rate & $3\text{e}^{-4}$  & $3\text{e}^{-4}$ & $3\text{e}^{-4}$ & $3\text{e}^{-4}$ & $3\text{e}^{-4}$ & $3\text{e}^{-4}$ & $3\text{e}^{-4}$ & $4\text{e}^{-4}$ \\
Steps & $5\text{e}^5$  & $5\text{e}^5$ & $5\text{e}^5$ & $2.5\text{e}^5$ & $2.5\text{e}^5$ & $5\text{e}^5$ & $5\text{e}^5$ & $2.5\text{e}^5$  \\
\midrule
\textbf{Hopper} & Diffuser & CG & S-Diff. & FM & SAD-F & SFM & CoBL & ASAP \\
\midrule
Batch Size & 128  & 128 & 128 & 128 & 128 & 128 & 128 & 128 \\
Learning Rate & $2\text{e}^{-4}$  & $2\text{e}^{-4}$ & $2\text{e}^{-4}$ & $2\text{e}^{-4}$ & $2\text{e}^{-4}$ & $2\text{e}^{-4}$ & $2\text{e}^{-4}$ & $4\text{e}^{-4}$ \\
Steps & $2.5\text{e}^5$  & $2.5\text{e}^5$ & $2.5\text{e}^5$ & $2.5\text{e}^5$ & $2.5\text{e}^5$ & $2.5\text{e}^5$ & $2.5\text{e}^5$ & $2.5\text{e}^5$ \\
\midrule
\textbf{KUKA Block-Stacking} & Diffuser & CG & S-Diff. & FM & SAD-F & SFM & CoBL & ASAP \\
\midrule
Batch Size & 64 & 64 & 64 & 64 & 64 & 64 & 64 & 64 \\
Learning Rate & $2\text{e}^{-5}$ & $2\text{e}^{-5}$ & $2\text{e}^{-5}$ & $2\text{e}^{-4}$ & $2\text{e}^{-4}$ & $2\text{e}^{-5}$ & $2\text{e}^{-5}$  & $1\text{e}^{-4}$ \\
Steps & $7\text{e}^5$ & $7\text{e}^5$ & $7\text{e}^5$ & $7\text{e}^5$ & $7\text{e}^5$ & $7\text{e}^5$ & $7\text{e}^5$  & $7\text{e}^5$ \\
\midrule
\textbf{Adroit Relocate} & Diffuser & CG & S-Diff. & FM & SAD-F & SFM & CoBL & ASAP \\
\midrule
Batch Size & 64 & 64 & 64 & 64 & 64 & 128 & 64 & 128 \\
Learning Rate & $2\text{e}^{-5}$ & $2\text{e}^{-5}$ & $2\text{e}^{-4}$ & $2\text{e}^{-4}$ & $2\text{e}^{-4}$ & $2\text{e}^{-4}$ & $2\text{e}^{-4}$  & $2\text{e}^{-4}$ \\
Steps & $1\text{e}^6$ & $1\text{e}^6$ & $1\text{e}^6$ & $1\text{e}^6$ & $1\text{e}^6$ & $2.5\text{e}^6$ & $1\text{e}^6$  & $5\text{e}^5$ \\
\bottomrule
\end{tabular}
\end{table}

%% file: tables/append_comp_resource.tex
\begin{table}[t]
\centering
\small
\caption{\textbf{Compute.} Hardware used for all training and evaluation for all experiments.}
\label{tab:append_comp}
\begin{tabular}{cccc}
\toprule
\textbf{Usage} & \textbf{CPU} & \textbf{GPU} & \textbf{GPU RAM} \\
\midrule
Simulation task training & AMD EPYC 7313P & NVIDIA RTX 4090 & 24 GB \\
Simulation task evaluation & AMD EPYC 7313P & NVIDIA RTX 4090 & 24 GB \\
Real-world task training & AMD EPYC 7313P & NVIDIA RTX 4090 & 24 GB 
\\
Real-world task evaluation & AMD Ryzen 7950X3D & NVIDIA RTX 4080 & 16 GB 
\\
\bottomrule
\end{tabular}
\end{table}